\documentclass[sigconf]{acmart}

\AtBeginDocument{%
  }

\copyrightyear{2025}
\acmYear{2025}
\setcopyright{acmlicensed}\acmConference[MM '25]{Proceedings of the 33rd ACM International Conference on Multimedia}{October 27--31, 2025}{Dublin, Ireland}
\acmBooktitle{Proceedings of the 33rd ACM International Conference on Multimedia (MM '25), October 27--31, 2025, Dublin, Ireland}
\acmDOI{10.1145/3746027.3754573}
\acmISBN{979-8-4007-2035-2/2025/10}

\usepackage{array}
\usepackage{multirow}
\usepackage{makecell}
\usepackage{adjustbox}
\usepackage[skip=4pt]{caption}
\usepackage{enumitem}
\usepackage{balance}
\usepackage{hyperref}

\begin{document}


\title{VideoForest: Person-Anchored Hierarchical Reasoning for Cross-Video Question Answering
}

\author{Yiran Meng}
\authornote{Both authors contributed equally to this research.}
\orcid{0009-0001-5856-2490}
\affiliation{%
  \institution{Sun Yat-Sen University}
  \city{Zhuhai}
  \country{China}
}

\author{Junhong Ye}
\authornotemark[1]
\affiliation{%
  \institution{Sun Yat-Sen University}
  \city{Zhuhai}
  \country{China}
}

\author{Wei Zhou}
\affiliation{%
  \institution{Cardiff University}
  \country{United Kingdom}
  }

\author{Guanghui Yue}
\affiliation{%
  \institution{Shenzhen University}
  \city{Shenzhen}
  \country{China}
}

\author{Xudong Mao}
\affiliation{%
 \institution{Sun Yat-Sen University}
 \city{Zhuhai}
 \country{China}
 }

\author{Ruomei Wang}
\affiliation{%
  \institution{Sun Yat-Sen University}
  \city{Guangzhou}
  \country{China}}

\author{Baoquan Zhao}
\orcid{0000-0002-0574-1663}
\authornote{Corresponding author. E-mail: zhaobaoquan@mail.sysu.edu.cn}
\affiliation{%
  \institution{Sun Yat-Sen University}
  \city{Zhuhai}
  \country{China}
}

\renewcommand{\shortauthors}{Yiran Meng et al.}


\begin{abstract}
Cross-video question answering presents significant challenges beyond traditional single-video understanding, particularly in establishing meaningful connections across video streams and managing the complexity of multi-source information retrieval. We introduce VideoForest, a novel framework that addresses these challenges through person-anchored hierarchical reasoning. Our approach leverages person-level features as natural bridge points between videos, enabling effective cross-video understanding without requiring end-to-end training. VideoForest integrates three key innovations: 1) a human-anchored feature extraction mechanism that employs ReID and tracking algorithms to establish robust spatiotemporal relationships across multiple video sources; 2) a multi-granularity spanning tree structure that hierarchically organizes visual content around person-level trajectories; and 3) a multi-agent reasoning framework that efficiently traverses this hierarchical structure to answer complex cross-video queries. To evaluate our approach, we develop CrossVideoQA, a comprehensive benchmark dataset specifically designed for person-centric cross-video analysis. Experimental results demonstrate VideoForest's superior performance in cross-video reasoning tasks, achieving 71.93\% accuracy in person recognition, 83.75\% in behavior analysis, and 51.67\% in summarization and reasoning, significantly outperforming existing methods. Our work establishes a new paradigm for cross-video understanding by unifying multiple video streams through person-level features, enabling sophisticated reasoning across distributed visual information while maintaining computational efficiency.

\end{abstract}







\begin{abstract}
Cross-video question answering presents significant challenges beyond traditional single-video understanding, particularly in establishing meaningful connections across video streams and managing the complexity of multi-source information retrieval. We introduce VideoForest, a novel framework that addresses these challenges through person-anchored hierarchical reasoning, enabling effective cross-video understanding without requiring end-to-end training. VideoForest integrates three key innovations: 1) a human-anchored feature extraction mechanism that employs ReID and tracking algorithms to establish robust spatiotemporal relationships across multiple video sources; 2) a multi-granularity spanning tree structure that hierarchically organizes visual content around person-level trajectories; and 3) a multi-agent reasoning framework that efficiently traverses this hierarchical structure to answer complex queries. To evaluate our method, we develop CrossVideoQA\footnote{The dataset can be reached via \href{https://github.com/liriar/VideoForest}{https://github.com/liriar/VideoForest}}, a comprehensive benchmark specifically designed for person-centric cross-video analysis. Experimental results demonstrate VideoForest's superior performance in cross-video reasoning tasks, achieving 71.93\% accuracy in person recognition, 83.75\% in behavior analysis, and 51.67\% in summarization and reasoning.

\end{abstract}


\maketitle

\section{Introduction}
\begin{sloppypar}

Cross-video understanding represents one of the most challenging frontiers in computer vision, requiring systems to extract, correlate, and reason about information distributed across multiple video streams. Unlike single-video analysis, where context remains contained within temporal boundaries, cross-video reasoning demands sophisticated mechanisms to establish meaningful connections across different spatial viewpoints and temporal sequences. This capability is particularly crucial in surveillance and monitoring scenarios, where critical information is inherently fragmented across multiple cameras, necessitating unified analysis for comprehensive situational awareness.

Consider a security investigation requiring analysts to determine: \textit{Which individual traversed all three campus buildings between 14:00-16:00?} Answering such queries demands not only person identification and tracking within each video stream but cross-referencing identities and behaviors across multiple cameras with varying perspectives and recording conditions. 
Despite remarkable advances in video understanding, current methods remain fundamentally constrained by their single-stream processing paradigm, rendering them inadequate for queries that span multiple video sources. This architectural limitation prevents the integration of complementary information across camera viewpoints.

\begin{figure}[!t]
    \centering
    \includegraphics[width=\linewidth]{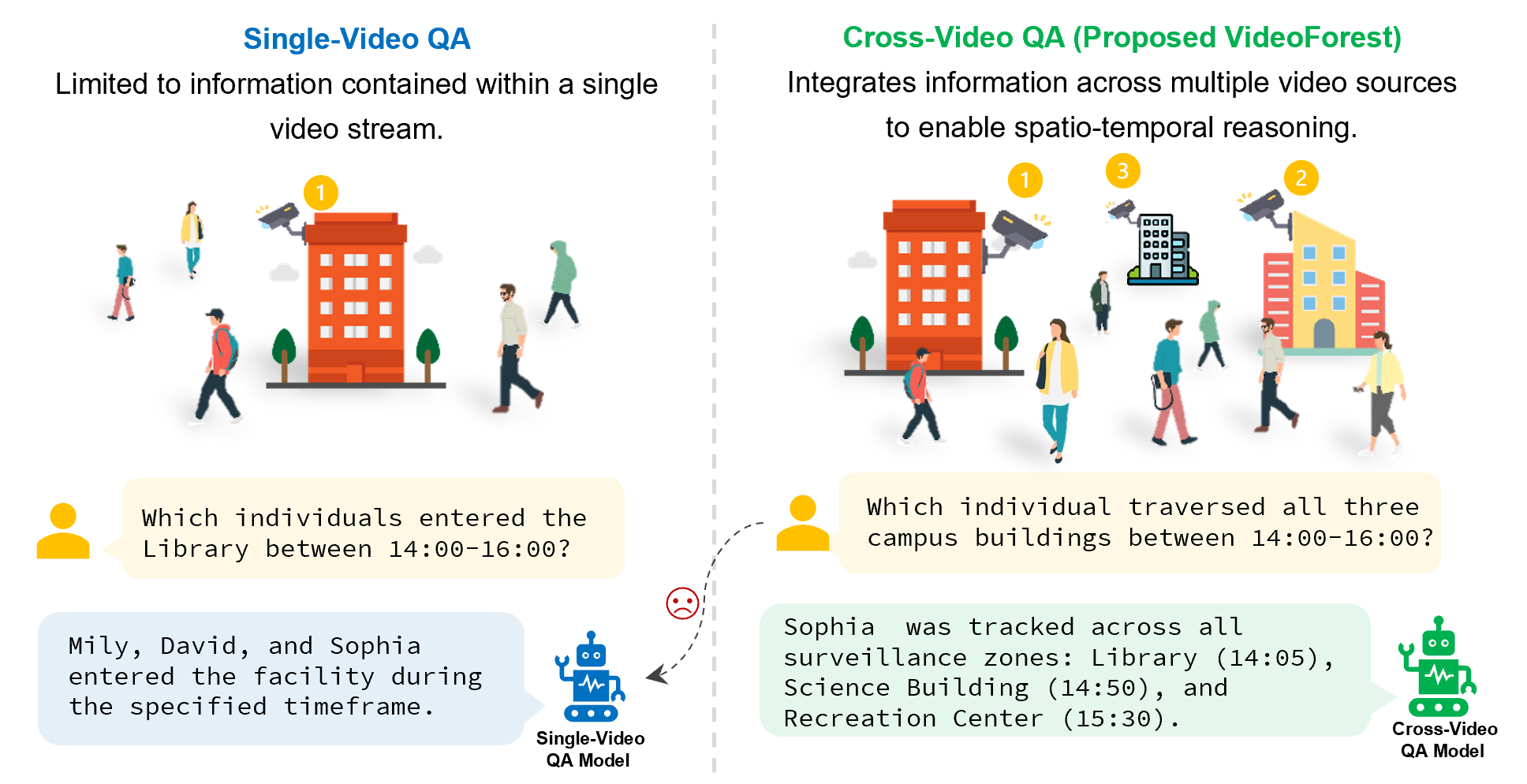}
    \caption{Comparison of single-video vs. cross-video question answering paradigms.}
    \label{fig:fig1}
\end{figure}

As illustrated in Figure~\ref{fig:fig1}, existing video question answering systems \cite{damonlpsg2023videollama,fan2025videoagent,Maaz2023VideoChatGPT,10656135} have primarily focused on maximizing performance within the bounds of individual videos, inadvertently reinforcing this single-stream constraint. Even recent advancements like VideoAgent \cite{fan2025videoagent}, which introduces sophisticated agent-based search strategies, and Chat-Video \cite{wang2023chatvideo}, which pioneers motion trajectory analysis, ultimately operate within the confines of isolated video processing. The critical capability of establishing semantic bridges between separate video streams—essential for true cross-video reasoning—remains largely unexplored in the literature.

\begin{figure*}[!t]
  \includegraphics[width=\textwidth]{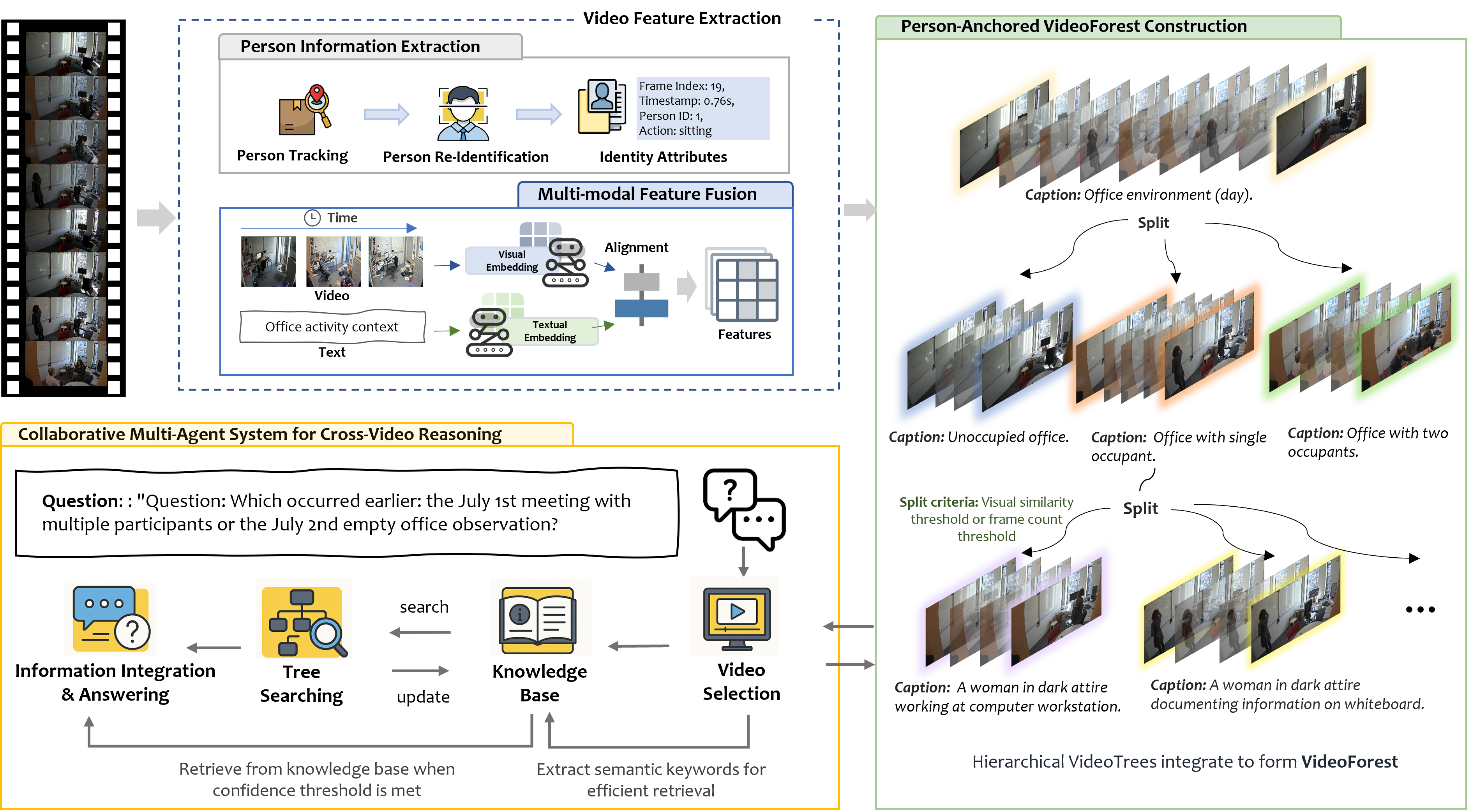}
\caption{VideoForest architecture for cross-video question answering.}
  \label{fig:teaser}
\end{figure*}

To address these limitations, we introduce \textbf{VideoForest}, a novel hierarchical framework that enables efficient person-centric reasoning across multiple video streams (see Figure~\ref{fig:teaser}). Our key insight is that human subjects serve as natural bridge points between different videos, providing consistent reference entities around which cross-video relationships can be structured. VideoForest implements this insight through three innovative components:
First, we develop a person-anchored feature extraction mechanism that employs Re-Identification (ReID) and tracking algorithms to establish consistent identity representations across multiple videos, creating robust spatio-temporal relationships that span different camera viewpoints. Second, we design a multi-granularity spanning tree structure that hierarchically organizes visual content around person-level trajectories, enabling efficient navigation from coarse scene-level information to fine-grained behavioral details. Third, we implement a multi-agent reasoning framework that efficiently traverses this hierarchical structure to perform sophisticated cross-video reasoning while maintaining computational tractability.

To evaluate our approach and advance research in cross-video understanding, we introduce \textbf{CrossVideoQA}, the first comprehensive benchmark dataset specifically designed for person-centric cross-video question answering in surveillance scenarios. Our extensive experiments demonstrate VideoForest's effectiveness across various reasoning tasks.

The primary contributions of this work are threefold:
\begin{itemize}
    \item We introduce the first person-anchored hierarchical framework for cross-video question answering, pioneering a tree-based architecture that uses human subjects as bridge points to connect multiple video streams, enabling unified understanding across distributed visual information.
    
    \item We develop an efficient multi-granular video organization strategy integrated with a multi-agent reasoning framework that preserves critical temporal-spatial relationships while making cross-video question answering computationally tractable.
    
    \item We present CrossVideoQA, a novel benchmark dataset for evaluating person-centric cross-video question answering capabilities, establishing new evaluation protocols and performance baselines for this emerging research direction.
\end{itemize}

\section{Related Work}

\subsection{Video Question Answering}

VideoQA forms a cornerstone of multimodal comprehension alongside text-video retrieval and video captioning, demanding deep understanding of complex semantic and causal relationships between video and language~\cite{liu2025semantic}.
To improve model robustness and interpretability, visual localization approaches enable models to highlight relevant video segments~\cite{10.1007/978-3-031-72652-1_6,liu2025commonsensevideoquestionanswering} or keyframes~\cite{yu2023selfchainedimagelanguagemodelvideo,wang2024videotree} for answer generation. While these methods successfully locate evidence, the reasoning process remains opaque. ChatVideo~\cite{wang2023chatvideo} introduces motion trajectories as fundamental analysis units, leveraging specialized visual foundation models to generate attribute annotations for enhanced temporal modeling in dynamic scenes.
Another research line employs external Large Language Models (LLMs) as reasoning modules. LLoVi~\cite{zhang2024simplellmframeworklongrange} converts Video QA into text-based QA via video captions, while VideoAgent~\cite{fan2025videoagent} recursively assesses frame sufficiency for question answering. Although improving QA performance, these methods depend heavily on LLM linguistic reasoning, which is prone to hallucination and lacks interpretability. Video-CCAM~\cite{fei2024videoccamenhancingvideolanguageunderstanding} introduces cross-modal attention with causal masking between visual encoders and LLMs, demonstrating superior performance across multiple benchmarks. InternVL 2.5~\cite{wang2025internvideo,li2024videochatflash} refines training strategies and data quality while excelling in short-video understanding, long-video retrieval, and QA tasks.
TV-trees~\cite{sanders2024tvtreesmultimodalentailmenttrees} adopts neuro-symbolic approaches for explicit reasoning across visual and textual modalities, though assuming pre-transcribed video text. Despite advances in single-video tasks, systems capable of cross-video understanding and complex temporal QA remain scarce.

\subsection{Structured Video Representation}
Structured video representation enhances Video QA by transforming videos from frame sequences into hierarchical semantic representations, enabling better comprehension of object, action, and event relationships \cite{xiao2022video}. This approach fuses visual information at multiple granularities with corresponding linguistic concepts, improving accuracy and interpretability \cite{Wang_2021_ICCV,10508488,10.1145/3581783.3612838}.
Recent video-language methods emphasize structured frame representations for efficient scene understanding \cite{Park2024TooMF,islam2024video,wang2024videotree,cheng2024enhancinglongvideounderstanding,ma2024drvideodocumentretrievalbased}. LVNet \cite{Park2024TooMF} reduces redundancy through hierarchical keyframe selection, VideoReCap \cite{islam2024video} introduces progressive captioning bridging short and long clip comprehension, and VideoTree \cite{wang2024videotree} achieves breakthroughs with top-down video language embedding featuring dynamic depth adjustment for efficient long video comprehension.
However, these approaches primarily address single-video analysis, leaving cross-video correlation challenges unexplored. Our work extends these foundational ideas to multi-video domains by introducing human-centered connectivity and spatio-temporal relationship modeling for effective cross-video understanding.

\subsection{Video Understanding Benchmarks}
Video understanding tasks progress through three complexity levels: abstract understanding for overall video events \cite{carreira2022shortnotekinetics700human,Heilbron2015ActivityNetAL}, temporal understanding for specific moment identification \cite{sigurdsson2016hollywoodhomescrowdsourcingdata,liu2022fineactionfinegrainedvideodataset}, and spatio-temporal understanding for timing and spatial localization \cite{benshabat2023ikeaasmdatasetunderstanding,Zhou_2023}. This progression mirrors human cognitive processes of building comprehensive understanding from visual information \cite{madan2024foundationmodelsvideounderstanding}.
Multimodal large language models (MLLMs) have catalyzed comprehensive evaluation frameworks \cite{fu2024videoMMEBenchmark,Li2024VideoVistaAV,li2024mvbench,mangalam2023egoschema,ning2023video}. Video-MM \cite{fu2024videoMMEBenchmark} established the first holistic MLLM evaluation benchmark, while VideoVistaAV \cite{Li2024VideoVistaAV} introduced multifaceted evaluation accounting for diverse content categories, time scales, and inference capabilities.
Despite these advances, existing benchmarks focus predominantly on single-video comprehension, creating a critical gap in cross-video reasoning assessment. Our proposed CrossVideoQA benchmark addresses this limitation through carefully curated cross-video queries based on Edinburgh Office Surveillance and HACS datasets \cite{qasim2021groundtruthing,zhao2019hacs}, establishing novel evaluation criteria for multi-video understanding systems.

\section{Methodology}

\subsection{Problem Definition and Notation}
We formalize the cross-video question answering task as follows. Let $\mathcal{V} = \{V_1, V_2, \ldots, V_n\}$ denote a collection of $n$ videos, where each video $V_i$ comprises an ordered temporal sequence of frames $\mathcal{F}_i = \{f_{i,1}, f_{i,2}, \ldots, f_{i,m_i}\}$ with $m_i$ denoting the number of frames in video $i$. Our objective is to construct a unified hierarchical representation that enables efficient cross-video information retrieval and reasoning.

For each frame $f_{i,j}$ (the $j$-th frame of video $i$), we extract two complementary representations: (1) visual embeddings $\mathbf{v}(f_{i,j}) \in \mathbb{R}^d$, a $d$-dimensional dense representation capturing semantic visual content; and (2) person detections $\mathbf{p}(f_{i,j}) = \{(t_k, \mathbf{x}_k, \textrm{id}_k)\}_{k=1}^{K_{i,j}}$, where $K_{i,j}$ is the number of detected persons in frame $f_{i,j}$, $t_k \in \mathbb{R}$ denotes the timestamp, $\mathbf{x}_k = (x_k, y_k) \in \mathbb{R}^2$ represents spatial coordinates, and $\textrm{id}_k \in \mathcal{I}$ is the unique person identifier from the set of all identities $\mathcal{I}$.

The cross-video question answering task can then be formally defined as a mapping function:
\begin{equation}
\mathcal{Q}: \mathcal{V} \times \mathcal{T} \times \mathcal{L} \rightarrow \mathcal{A},
\end{equation}
where $\mathcal{T} \subset \mathbb{R}^+$ represents temporal constraints (e.g., time intervals of interest), $\mathcal{L} \subset \mathbb{R}^2$ represents spatial constraints (e.g., regions of interest), and $\mathcal{A}$ denotes the answer space, which may include textual responses, temporal localization, or entity identification. This formulation explicitly models the cross-video reasoning process as conditional on both temporal and spatial constraints, capturing the complex spatio-temporal relationships inherent in surveillance and monitoring scenarios. Our hierarchical VideoForest framework implements this mapping through a person-anchored tree structure that enables efficient traversal and integration of information across multiple video sources.

\subsection{Dual-Stream Feature Extraction and Adaptive Segmentation}
Building on our formal problem definition, we implement a complementary dual-stream architecture for comprehensive video representation. The visual content stream employs the ViCLIP encoder~\cite{wang2023internvid,wang2022internvideo}, parameterized by $\theta_v$, to compute the frame-level embeddings defined in our notation:
\begin{equation}
\mathbf{v}(f_{i,j}) = \phi(f_{i,j}; \theta_v) \in \mathbb{R}^d.
\end{equation}

Concurrently, the person-centric stream utilizes a specialized tracking model $\psi$ with parameters $\theta_p$ to identify and extract the structured person representations:
\begin{equation}
\mathbf{p}(f_{i,j}) = \psi(f_{i,j}; \theta_p) = \{(t_k, \mathbf{x}_k, \textrm{id}_k)\}_{k=1}^{K_{i,j}},
\end{equation}
where $K_{i,j}$ denotes the number of persons detected in frame $f_{i,j}$.

To partition videos into semantically coherent segments, we define an adaptive boundary detection function $S: \mathcal{F}_i \rightarrow \{0,1\}$ that identifies significant transitions through a disjunctive criterion:

\begin{equation}
S(f_{i,j}) = \mathbb{1}[C_1(f_{i,j}) \vee C_2(f_{i,j}) \vee C_3(f_{i,j})],
\end{equation}
where the three complementary criteria are formulated as:

\begin{align}
C_1(f_{i,j}): \|\mathbf{v}(f_{i,j}) - \mathbf{v}(f_{i,j+1})\|_2 &> \epsilon_1, \quad \text{(local transition)} \\
C_2(f_{i,j}): \|\mathbf{v}(f_{i,j}) - \mathbf{v}(f_{i,j}^{\text{cent}})\|_2 &> \epsilon_2, \quad \text{(global deviation)} \\
C_3(f_{i,j}): |\mathcal{P}(f_{i,j}) \triangle \mathcal{P}(f_{i,j-1})| &\geq \Delta_{\mathcal{P}}. \quad \text{(person-set change)}
\end{align}

Here, $\mathbb{1}[\cdot]$ denotes the indicator function, $\mathcal{P}(f_{i,j}) = \{\textrm{id}_k | (t_k, \mathbf{x}_k, \textrm{id}_k) \in \mathbf{p}(f_{i,j})\}$ represents the set of person identities present in frame $f_{i,j}$, and $f_{i,j}^{\text{cent}}$ refers to a representative frame for the current segment. The multi-criterion approach operates at three distinct levels: $C_1$ captures frame-to-frame appearance changes through local feature distance, $C_2$ measures deviation from the segment's visual prototype to identify global content drift, and $C_3$ quantifies person-centric dynamics through the cardinality of the symmetric difference $\triangle$ between consecutive person-identity sets. 
The thresholds $\epsilon_1$, $\epsilon_2$, and $\Delta_{\mathcal{P}}$ are determined through cross-validation on a held-out dataset to optimize the trade-off between temporal granularity and semantic coherence. When a segment boundary is detected according to $S(f_{i,j})=1$, we create a new segment $S_{i,k} = \{f_{i,j_{\text{start}}}, f_{i,j_{\text{start}}+1}, \ldots, f_{i,j_{\text{end}}}\}$, where $j_{\text{start}}$ and $j_{\text{end}}$ denote the inclusive boundaries of the segment. This approach yields a sequence of non-overlapping segments $\{S_{i,1}, S_{i,2}, \ldots, S_{i,n_i}\}$ for each video $V_i$, effectively parsing the continuous video stream into discrete semantic units.

This adaptive segmentation serves as the foundational building block for our hierarchical tree representation, enabling efficient multi-granular video indexing and retrieval. The segments preserve semantic coherence while establishing manageable units for subsequent person-anchored correlation across videos. By incorporating both visual content and person-centric dynamics in the segmentation criteria, our approach ensures that the resulting segments maintain meaningful contextual boundaries that facilitate cross-video reasoning.

\subsection{Multi-Level Semantic Representation}
Given the segmented video structure, we construct semantically rich representations for each segment $S_{i,k}$. We define a multi-modal encoding function $\eta: \mathcal{S} \times \mathcal{P} \rightarrow \mathbb{R}^d$ that maps visual content and person trajectories to a unified semantic space:

\begin{equation}
\mathbf{C}(S_{i,k}) = \eta(\mathbf{v}(f_{i,j}^{\text{key}}), \mathbf{P}(S_{i,k}); \theta_{\eta}),
\end{equation}
where $f_{i,j}^{\text{key}} = f_{i,\lfloor(j_{\text{start}} + j_{\text{end}})/2\rfloor}$ is the temporally central keyframe representing the segment, $\mathbf{P}(S_{i,k}) = \{\mathbf{p}(f_{i,j}) | f_{i,j} \in S_{i,k}\}$ denotes the aggregated set of person detections across all frames in the segment, and $\theta_{\eta}$ parameterizes the encoding function. This formulation ensures that our semantic representation captures both the static visual content through the keyframe embedding and the dynamic person-centric activities through trajectory aggregation.

This multi-level semantic representation provides a rich foundation for cross-video reasoning by capturing both visual scene context and person-centric dynamics. The resulting segment-level encodings $\{\mathbf{C}(S_{1,1}), \mathbf{C}(S_{1,2}), \ldots, \mathbf{C}(S_{n,n_n})\}$ serve as the semantic nodes in our hierarchical tree structure, enabling efficient retrieval and correlation of content across multiple videos.

\begin{figure*}[t]
    \centering
    \includegraphics[width=\linewidth]{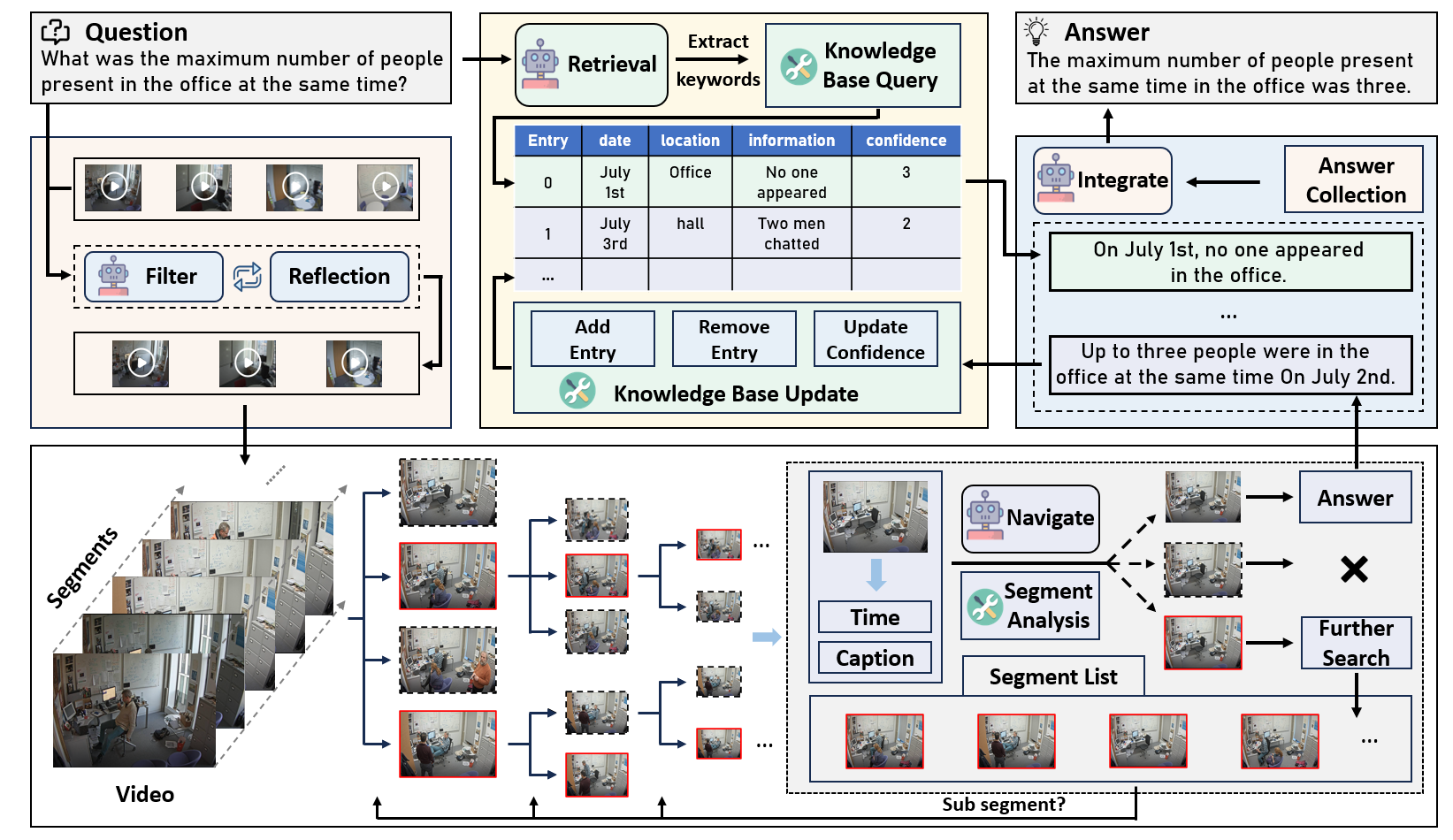}
    \caption{Architecture of our distributed multi-agent framework for cross-video reasoning.
    }
    \label{fig:query}
\end{figure*} 

\subsection{VideoForest Construction}

Based on the segmented videos and their semantic representations, we construct a hierarchical tree structure $\mathcal{T} = (V, E)$ that organizes content at multiple granularities. Each node $v \in V$ is defined as a structured tuple:
\begin{equation}
v = (t_{\text{start}}, t_{\text{end}}, \mathcal{R}_v, \mathbf{C}_v, \Gamma_v),
\end{equation}
where $[t_{\text{start}}, t_{\text{end}}] \subset \mathbb{R}^+$ represents the temporal interval spanned by the node, $\mathcal{R}_v = \{(\textrm{id}_k, \boldsymbol{\tau}_k)\}_{k=1}^{K_v}$ contains person re-identification information with $\textrm{id}_k \in \mathcal{I}$ and trajectory descriptors $\boldsymbol{\tau}_k$, $\mathbf{C}_v \in \mathbb{R}^d$ denotes the semantic content representation computed by function $\eta$, and $\Gamma_v \subset V$ represents the set of child nodes.

The edge set $E = \{(v_i, v_j) \mid v_j \in \Gamma_{v_i}\}$ defines the hierarchical parent-child relationships that enable efficient multi-granular traversal. To ensure comprehensive temporal coverage while maintaining non-overlapping child segments, the recursive partitioning of nodes follows a disjoint cover principle implemented by the splitting function $\text{Split}: V \rightarrow 2^V$:
\begin{equation}
\text{Split}(v) = \{v_1, v_2, \ldots, v_{K_v}\},
\end{equation}
satisfying the following temporal coverage and disjointness constraints:

\begin{equation}
\bigcup_{i=1}^{K_v} [t_{\text{start}}(v_i), t_{\text{end}}(v_i)] = [t_{\text{start}}(v), t_{\text{end}}(v)],
\end{equation}

\begin{equation}
\forall i \neq j: [t_{\text{start}}(v_i), t_{\text{end}}(v_i)] \cap [t_{\text{start}}(v_j), t_{\text{end}}(v_j)] = \emptyset.
\end{equation}

The splitting criteria are determined adaptively based on semantic similarity and person-level continuity, with splitting boundaries preferentially aligned with the segment boundaries identified during the segmentation process. For each video $V_i$, we construct a corresponding tree $\mathcal{T}_i$ with the root node spanning the entire video duration and leaf nodes corresponding to the fine-grained segments $\{S_{i,k}\}_{k=1}^{n_i}$.
This hierarchical organization enables efficient top-down traversal from coarse temporal chunks to fine-grained segments, facilitating rapid identification of relevant content in response to temporal and person-centric queries. The integration of person re-identification information $\mathcal{R}_v$ at each node creates natural bridge points between different video trees, enabling cross-video correlation and reasoning based on person identity continuity.

\subsection{Collaborative Multi-Agent System for Cross-Video Reasoning}

As illustrated in Fig.~\ref{fig:query}, our cross-video reasoning system integrates information from multiple video sources through a coordinated multi-agent architecture. This approach addresses the challenges of spatio-temporal relationships between videos, such as different viewpoints of the same scene or recordings from the same viewpoint at different times. The system employs four specialized agent modules working in concert to facilitate efficient cross-video reasoning.
Our multi-agent reasoning system is implemented using the CrewAI framework~\cite{crewai2024}, which provides modular agent-task scheduling and tool integration capabilities. We extend CrewAI to support dynamic tree-based traversal and cross-agent knowledge propagation.

\subsubsection{Agent Architecture and Functional Specialization}
Our multi-agent framework consists of four specialized components, each with distinct functionality:
\begin{equation}
\mathcal{A} = \{\mathcal{A}_{\text{filter}}, \mathcal{A}_{\text{retrieval}}, \mathcal{A}_{\text{navigate}}, \mathcal{A}_{\text{integrate}}\}
\end{equation}
The $\mathcal{A}_{\text{filter}}$ agent processes input queries to extract temporal and spatial constraints, identifying and selecting relevant video tree structures from the set $\{\mathcal{T}_i\}_{i=1}^n$. The $\mathcal{A}_{\text{retrieval}}$ agent manages knowledge base access, retrieving pertinent information while preventing redundant computations through confidence-based retrieval mechanisms. The $\mathcal{A}_{\text{navigate}}$ agent traverses the hierarchical tree structures using an optimized search strategy to locate query-relevant information. Finally, the $\mathcal{A}_{\text{integrate}}$ agent synthesizes information from both the knowledge base and tree structures, performing cross-video reasoning to generate comprehensive answers.

The reasoning workflow follows a sequential five-stage process: (1) video selection, (2) knowledge base retrieval, (3) hierarchical tree traversal, (4) cross-video information integration, and (5) knowledge base updating.

\subsubsection{Knowledge Base Construction and Confidence-Based Maintenance}
To address the computational challenges of repeatedly accessing the same information across multiple queries, we implement a global knowledge base $\mathcal{K}$ with confidence-weighted entries:
\begin{equation}
\mathcal{K} = \{(d_i, l_i, s_i, c_i)\}_{i=1}^N,
\end{equation}
where $d_i \in \mathcal{D}$ represents date information, $l_i \in \mathcal{L}$ denotes spatial location, $s_i \in \mathcal{S}$ is a descriptive string containing subject and action information, and $c_i \in [0, C_{\text{max}}]$ is the confidence score.

The $\mathcal{A}_{\text{retrieval}}$ agent maintains the integrity of $\mathcal{K}$ through a formal update function $\mathcal{U}: \mathcal{K}_{\text{new}} \times \mathcal{K} \rightarrow \mathcal{K}$ defined as:

\begin{equation}
\resizebox{\hsize}{!}{$
\mathcal{U}(k_{\text{new}}, \mathcal{K}) = 
\begin{cases}
\mathcal{K} \cup \{(d_{\text{new}}, l_{\text{new}}, s_{\text{new}}, 1)\}, & \text{if } k_{\text{new}} \notin \mathcal{K} \\
\mathcal{K} \setminus \{k_i\} \cup \{(d_i, l_i, s_i, c_i+1)\}, & \text{if } k_{\text{new}} = k_i \in \mathcal{K} \\
\mathcal{K} \setminus \{k_i\} \cup \{(d_i, l_i, s_i, c_i-1)\} & \text{if }, k_{\text{new}} \approx k_i \text{ and } c_i > 2 \\
\mathcal{K} \setminus \{k_i\} \cup \{(d_{\text{new}}, l_{\text{new}}, s_{\text{new}}, 1)\}, & \text{if } k_{\text{new}} \approx k_i \text{ and } c_i \leq 2
\end{cases}$
}
\end{equation}
where $k_{\text{new}} \approx k_i$ denotes a semantic conflict between the new entry and an existing entry according to a similarity measure $\text{sim}(k_{\text{new}}, k_i) > \tau_{\text{sim}}$.

This confidence-based approach enables the system to self-correct over time, progressively refining the knowledge base through iterative query answering. When processing new queries, entries with confidence scores exceeding a threshold $\tau_{\text{conf}}$ are prioritized for retrieval, reducing computational load and improving response time.

\subsubsection{Adaptive Hierarchical Search Optimization}
The $\mathcal{A}_{\text{navigate}}$ agent employs an efficient top-down search strategy $\mathcal{S}: \mathcal{Q} \times V \rightarrow 2^{\mathbf{C}}$ that recursively explores the hierarchical structure. For a query $q \in \mathcal{Q}$ and a node $v \in V$ within tree $\mathcal{T}$, the search function is formulated as:

\begin{equation}
\mathcal{S}(q, v) = 
\begin{cases}
\mathbf{C}_v, & \text{if } \text{Relevance}(q, \mathbf{C}_v) \geq \tau_{\text{rel}} \\
\bigcup\limits_{v_c \in \Gamma_v} \mathcal{S}(q, v_c), & \text{otherwise}
\end{cases}
\end{equation}
where $\text{Relevance}: \mathcal{Q} \times \mathbb{R}^d \rightarrow [0,1]$ measures the semantic similarity between the query $q$ and the content representation $\mathbf{C}_v$ of node $v$, and $\tau_{\text{rel}} \in [0,1]$ is a configurable relevance threshold.

The search process begins at the root nodes of selected video trees and progressively refines the exploration based on temporal, spatial, and person-centric constraints extracted from the query. When person-level information is present in the query, the search leverages the ReID information $\mathcal{R}_v$ stored at each node to efficiently identify relevant content across different videos.


\section{Experimental Evaluation}

To comprehensively evaluate VideoForest's capabilities for cross-video reasoning, we require a benchmark specifically designed to test integration and understanding across multiple video sources with varying spatial and temporal relationships. Existing video QA benchmarks primarily focus on single-video understanding, making them inadequate for assessing cross-video reasoning performance. We first introduce our CrossVideoQA benchmark, then present implementation details and comparative results that demonstrate the effectiveness of our approach across multiple reasoning tasks and evaluation configurations.

\subsection{CrossVideoQA Benchmark}

We introduce \textbf{CrossVideoQA}, a comprehensive benchmark dataset specifically designed to evaluate cross-video reasoning capabilities. This benchmark addresses the fundamental challenges of integrating information across multiple video sources, with particular focus on human-centric queries that span different spatial locations and temporal periods.

\subsubsection{Dataset Construction}
Cross-video understanding has critical applications across multiple domains. In surveillance contexts, it enables tracking individuals across distributed camera networks in complex environments such as office buildings and transportation hubs. In content analysis scenarios, it facilitates the discovery of related events across different video perspectives, enabling comprehensive story reconstruction.
To support rigorous evaluation of these capabilities, CrossVideoQA integrates two complementary high-quality datasets:

\begin{equation}
\mathcal{D}_{\text{CrossVideoQA}} = \mathcal{D}_{\text{EOSD}} \cup \mathcal{D}_{\text{HACS}}
\end{equation}

The Edinburgh Office Surveillance Dataset~\cite{qasim2021groundtruthing} provides 18 surveillance videos captured across 3 distinct locations over 12 different dates, encompassing approximately 450,000 frames. This dataset is particularly valuable for analyzing structured human behavior patterns in controlled indoor environments. The HACS dataset~\cite{zhao2019hacs} contributes 50,000 videos containing 1.55 million action clips, offering greater diversity in action categories and environmental contexts.

\subsubsection{Evaluation Framework}
CrossVideoQA is structured around three progressively complex reasoning tasks that evaluate distinct aspects of cross-video understanding:

\begin{itemize}
\item \textbf{Person Recognition:} Evaluates a system's ability to identify and track specific individuals across multiple video sources, establishing person-level correspondence across spatial and temporal boundaries.

\item \textbf{Behavior Analysis:} Assesses the interpretation of human activities, interactions, and behavioral patterns that may span multiple videos, requiring integration of contextual information across sources.

\item \textbf{Summarization and Reasoning:} Tests advanced capabilities in synthesizing causal relationships, extracting insights, and performing logical inference across multiple videos to answer complex queries.
\end{itemize}

To comprehensively assess cross-video reasoning across different spatio-temporal configurations, we define four evaluation modalities that systematically increase in complexity:

\begin{equation}
\mathcal{M} = \{\mathcal{M}_{\text{single}}, \mathcal{M}_{\text{cross-spatial}}, \mathcal{M}_{\text{cross-temporal}}, \mathcal{M}_{\text{cross-spatiotemporal}}\}
\end{equation}
where $\mathcal{M}_{\text{single}}$ evaluates retrieval within a single video (same day, same location), $\mathcal{M}_{\text{cross-spatial}}$ requires integration across locations within the same time period, $\mathcal{M}_{\text{cross-temporal}}$ assesses temporal reasoning at a fixed location, and $\mathcal{M}_{\text{cross-spatiotemporal}}$ represents the most challenging scenario requiring full spatio-temporal integration.
This structured framework provides a comprehensive assessment of a system's cross-video understanding capabilities across a spectrum of increasingly complex scenarios, enabling targeted identification of both strengths and limitations.

\subsubsection{Benchmark Construction Methodology}
To ensure benchmark quality and relevance, we employed a rigorous three-phase question generation pipeline. First, domain specialists manually created high-quality exemplar questions for each reasoning category and evaluation modality, establishing gold-standard references. Second, large language models were employed to systematically augment the question set under constrained generation parameters, ensuring coverage and diversity. Finally, all generated questions underwent expert review to validate answerable status, factual accuracy, and appropriate difficulty calibration. This methodical approach yielded a diverse and challenging benchmark that systematically explores the capabilities required for effective cross-video reasoning.

\subsection{Implementation Details}
We conducted comprehensive experiments to evaluate VideoForest on the CrossVideoQA benchmark. All experiments were performed on an NVIDIA RTX 4090 GPU using PyTorch framework. For fair comparison, we implemented consistent evaluation protocols across all models.

\subsection{Comparative Methods}
We benchmarked VideoForest against state-of-the-art video understanding models:

\begin{itemize}[itemsep=0.3em, topsep=2pt, parsep=0pt]

\item \textbf{Video-CCAM}~\cite{fei2024videoccamenhancingvideolanguageunderstanding}: Incorporates a criss-cross attention mechanism with causal masking between visual encoder and language model, demonstrating strong performance across diverse video length domains.

\item \textbf{InternVL 2.5}~\cite{wang2025internvideo,li2024videochatflash}: An advanced multimodal language model that extends InternVL 2.0 with enhanced training strategies and data quality optimizations. 

\item \textbf{LLaVA-OneVision}~\cite{li2024llavaonevision,xiong2024llavacritic}: A unified model designed for cross-modal transfer learning across single-image, multi-image, and video understanding tasks.

\item \textbf{ShareGPT4Video}~\cite{chen2024sharegpt4video}: A video-language model trained on 4.8M high-quality videos, achieving state-of-the-art performance on multiple video understanding benchmarks. This model also provides the captioning component in our VideoForest framework.

\end{itemize}

For fair evaluation of these single-video models on cross-video tasks, we implemented a sequential processing protocol with explicit instructions to: (1) assess video relevance to the query, (2) extract pertinent information from relevant videos, and (3) synthesize extracted information into a coherent response.


\subsection{Performance Analysis}

\subsubsection{Task-Specific Performance Analysis}

Table~\ref{tab:performance_comparison} presents a comprehensive evaluation across the three fundamental reasoning categories in our benchmark. VideoForest demonstrates statistically significant performance advantages across all evaluation dimensions, with particularly pronounced improvements in Person Identification (+13.00\% compared to the strongest baseline) and Behavioral Analysis (+17.08\%). These consistent performance differentials validate the efficacy of our hierarchical person-anchored reasoning architecture in addressing cross-video understanding challenges.
Particularly noteworthy is the observation that even state-of-the-art single-video models such as InternVL-2.5, which achieve competitive results in traditional VideoQA tasks, exhibit substantial performance degradation in cross-video person recognition scenarios (58.93\% versus our 71.93\%). This performance gap underscores the critical importance of explicit person-level tracking and re-identification components for maintaining identity coherence across temporally and spatially distributed video segments. The results provide compelling evidence that VideoForest's multi-level reasoning approach effectively addresses the fundamental challenges in cross-video understanding.

\begin{table}[!t]
\centering

\caption{Quantitative performance comparison across the three reasoning categories defined in the CrossVideoQA benchmark.}
\renewcommand{\arraystretch}{1.0}
\setlength{\tabcolsep}{0.9pt}
  \resizebox{\linewidth}{!}{ 
\begin{tabular}{l|c|c|c|c}
\hline
\textbf{Model} & Person Rec. & Behav. Ana. & Sum. and Reas. & \textbf{Overall Acc.} \\
\hline
\makecell[l]{\textbf{ShareGPT4Video-8B}~\cite{chen2024sharegpt4video}} & 50.00 & 49.38 & 41.67 & 47.21 \\
\makecell[l]{\textbf{VideoCCAM-7B}~\cite{fei2024videoccamenhancingvideolanguageunderstanding}} & 42.86 & 41.98 & 45.00 & 43.15 \\
\makecell[l]{\textbf{InternVL-2.5}~\cite{wang2025internvideo}} & 58.93 & 66.67 & 46.67 & 58.38 \\
\makecell[l]{\textbf{LLaVAOneVision}~\cite{li2024llavaonevision}} & 51.79 & 53.09 & 36.67 & 47.72 \\
\makecell[l]{\textbf{ChatUniVi}~\cite{wang2023chatvideo}} & 46.43 & 66.67 & 30.00 & 49.75 \\
\makecell[l]{\textbf{LLaVA-NeXTVideo-7B}~\cite{zhang2024llavanextvideo,liu2024llavanext}} & 51.79 & 56.79 & 36.67 & 49.24 \\
\makecell[l]{\textbf{VideoChatFlash}~\cite{li2024videochatflash}} & 46.43 & 46.91 & 46.67 & 46.70 \\
\makecell[l]{\textbf{VideoLLaMA3-7B}~\cite{damonlpsg2025videollama3,damonlpsg2024videollama2,damonlpsg2023videollama}} & 46.43 & 50.62 & 33.33 & 44.16 \\
\makecell[l]{\textbf{LongVA-7B}~\cite{wu2024longvideobenchbenchmarklongcontextinterleaved}} & 44.64 & 64.20 & 38.33 & 50.76 \\
\makecell[l]{\textbf{BIMBA-LLaVA}~\cite{islam2025bimba}} & 57.14 & 71.60 & 36.67 & 58.38 \\
\makecell[l]{\textbf{mPLUG-Owl3}~\cite{ye2024mplugowl3longimagesequenceunderstanding}} & 57.79 & 71.60 & 38.33 & 55.84 \\
\hline
\makecell[l]{\textbf{VideoForest(Ours)}} & \textbf{71.93} & \textbf{83.75} & \textbf{51.67} & \textbf{69.12} \\
\hline
\end{tabular}
}
\label{tab:performance_comparison}
\end{table}

\begin{table}[!t]
\centering

\caption{Performance comparison across the four evaluation modalities in CrossVideoQA.}
\renewcommand{\arraystretch}{1.1}
\setlength{\tabcolsep}{1.2pt}
  \resizebox{\linewidth}{!}{ 
\begin{tabular}{l|c|c|c|c}
\hline
\textbf{Model} & $\mathcal{M}_{\text{cross-temporal}}$ & $\mathcal{M}_{\text{cross-spatial}}$ & $\mathcal{M}_{\text{cross-spa-tem}}$ & $\mathcal{M}_{\text{single}}$ \\
\hline
\makecell[l]{\textbf{ShareGPT4Video-8B}~\cite{chen2024sharegpt4video}} & 64.00 & 38.64 & 53.85 & 42.31 \\
\makecell[l]{\textbf{VideoCCAM-7B}~\cite{fei2024videoccamenhancingvideolanguageunderstanding}} & 60.00 & 46.15 & 50.00 & 57.69 \\
\makecell[l]{\textbf{InternVL-2.5}~\cite{wang2025internvideo}} & 52.00 & 46.15 & \textbf{69.23} & 73.08 \\
\makecell[l]{\textbf{LLaVAOneVision}~\cite{li2024llavaonevision}} & 36.00 & 61.54 & 53.85 & 50.00 \\
\makecell[l]{\textbf{ChatUniVi}~\cite{wang2023chatvideo}} & 64.00 & 38.46 & 38.46 & 57.69 \\
\makecell[l]{\textbf{LLaVA-NeXTVideo-7B}~\cite{zhang2024llavanextvideo,liu2024llavanext}} & 56.00 & 42.31 & 65.38 & 42.31 \\
\makecell[l]{\textbf{VideoChatFlash}~\cite{li2024videochatflash}} & 52.00 & 46.15 & 34.62 & 53.85 \\
\makecell[l]{\textbf{VideoLLaMA3-7B}~\cite{damonlpsg2025videollama3,damonlpsg2024videollama2,damonlpsg2023videollama}} & 48.00 & 46.15 & 42.31 & 46.15 \\
\makecell[l]{\textbf{LongVA-7B}~\cite{wu2024longvideobenchbenchmarklongcontextinterleaved}} & 52.00 & 50.00 & 34.62 & 50.00 \\
\makecell[l]{\textbf{BIMBA-LLaVA}~\cite{islam2025bimba}} & 48.00 & 46.15 &  \textbf{69.23}& 53.85 \\
\makecell[l]{\textbf{mPLUG-Owl3s}~\cite{ye2024mplugowl3longimagesequenceunderstanding}} & 68.00 & 46.15 &  65.38& 53.85 \\
\hline
\makecell[l]{\textbf{VideoForest(Ours)}} & \textbf{72.00} & \textbf{69.23} & 65.38 & \textbf{61.54} \\
\hline
\end{tabular}
}
\label{tab:module_analysis}
\end{table}

\begin{figure*}[t]
    \centering
    \includegraphics[width=\linewidth]{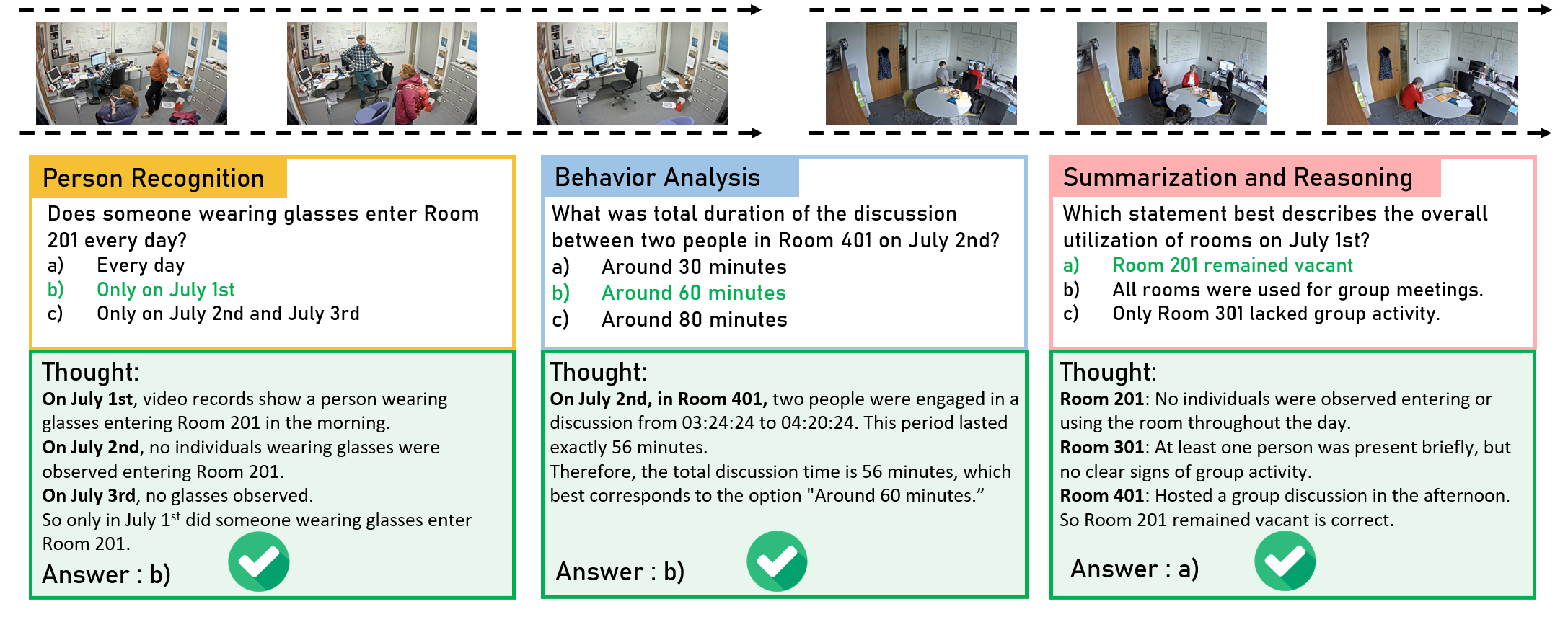}
    \caption{Exemplars from CrossVideoQA illustrating VideoForest's multi-modal reasoning architecture.}

    \label{fig:question examples and reasoning process}
\end{figure*} 

\subsubsection{Evaluation Across Spatio-Temporal Configurations}

Table~\ref{tab:module_analysis} presents performance analysis across four spatio-temporal configurations. VideoForest achieves 72.00\% accuracy on cross-temporal reasoning, outperforming ShareGPT4Video by 8.00\%, demonstrating our hierarchical tree structure's effectiveness in establishing temporal relationships. For cross-spatial integration, our model achieves 69.23\% accuracy, surpassing LLaVA-OneVision (61.54\%) by 7.69\%, validating our person-anchored approach for connecting information across spatial boundaries.
VideoForest maintains strong performance across all configurations, though the smallest gap occurs in cross-spatiotemporal tasks, indicating this remains challenging and suggesting future research directions. Existing models show specialization patterns—InternVL 2.5 excels in single-video tasks (73.08\%) but underperforms cross-temporally (52.00\%), while ShareGPT4Video shows opposite patterns. VideoForest demonstrates balanced performance across all configurations.

\subsection{Qualitative Analysis}

Figure~\ref{fig:question examples and reasoning process} presents examples of VideoForest's reasoning process and response generation. Our qualitative analysis reveals two key patterns: 1) VideoForest effectively employs two-stage reasoning for cross-video queries—first retrieving relevant information from individual video trees, then synthesizing coherent answers; 2) Primary failure modes involve fine-grained action recognition where the model cannot identify detailed actions like \textit{writing on paper} due to limited surveillance footage resolution, frame sampling constraints, and action ambiguity in complex environments.

\begin{table}[!tb]
\centering
\caption{Ablation study of VideoForest across four settings.}
\begin{adjustbox}{width=\linewidth}
\begin{tabular}{l|c|c|c}
\hline
\textbf{Setting} & \textbf{w/o Retrieval} & \textbf{w/o Reflection} & \textbf{Full Model} \\
\hline
$\mathcal{M}_{\text{cross-temporal}}$ & 60.00 & 48.00 & \textbf{72.00} \\
$\mathcal{M}_{\text{cross-spatial}}$ & 61.54 & 57.69 & \textbf{69.23} \\
$\mathcal{M}_{\text{cross-spa-tem}}$ & 50.00 & 46.15 & \textbf{65.38} \\
$\mathcal{M}_{\text{single}}$ & 50.00 & 42.31 & \textbf{61.54} \\
\hline
\textbf{Average} & 55.39 & 48.54 & \textbf{67.54} \\
\hline
\end{tabular}
\end{adjustbox}
\label{tab:ablation_study}
\end{table}

\begin{table}[!tb]
\centering
\caption{Ablation study on structural design choices in the video tree search module of VideoForest.}
\begin{adjustbox}{width=\linewidth}
\begin{tabular}{l|c|c|c|c}
\hline
\textbf{Setting} & \textbf{w/o ReID in Search} & \textbf{w/o Video Filter} & \textbf{w/o Deep Tree Traversal} & \textbf{Full Model}\\
\hline
$\mathcal{M}_{\text{cross-temporal}}$ & 52.00 & 44.00 & 60.00& \textbf{72.00} \\
$\mathcal{M}_{\text{cross-spatial}}$ & 53.85 & 61.54 &65.38& \textbf{69.23} \\
$\mathcal{M}_{\text{cross-spa-tem}}$ & 57.69 &38.46&61.54  & \textbf{65.38} \\
$\mathcal{M}_{\text{single}}$ & 50.00 & 61.54 &57.69& \textbf{61.54} \\
\hline
\textbf{Average} & 53.39 & 51.39&61.15 & \textbf{67.54} \\
\hline
\end{tabular}
\end{adjustbox}
\vspace{-1em}
\label{tab:ablation}
\end{table}
\subsection{Ablation Study}

We conducted ablation studies to quantify key component contributions in VideoForest. Table~\ref{tab:ablation_study} illustrates the performance impact when removing individual components.
Results demonstrate that each component contributes significantly to overall performance. Disabling knowledge base retrieval decreases performance by 10-25\%, particularly affecting cross spatial-temporal scenarios requiring context-dependent and precise knowledge. Eliminating the reflection component impacts spatio-temporal reasoning with the most significant drop ($\approx$ 33\%) in cross-temporal scenarios. These results validate our architectural design and highlight component complementarity for effective cross-video reasoning.
We conducted quantitative ablation studies on three key modules corresponding to core mechanisms in the video tree search process:
\begin{itemize}
    \item \textbf{w/o ReID in Search:} Person trajectories construct tree structure, but person IDs are not used for node filtering during inference, evaluating ReID impact as anchor signals.
    \item \textbf{w/o Video Filter:} All videos pass directly to downstream agents without pre-filtering based on question content.
    \item \textbf{w/o Deep Tree Traversal:} Only first-layer nodes (coarse-grained summaries) are retained, excluding deeper fine-grained details.
\end{itemize}
Table~\ref{tab:ablation} presents results for these three mechanisms. Removing any module causes consistent accuracy drops, demonstrating their indispensable roles. Disabling ReID-based filtering leads to a 20.00\% drop in cross-temporal settings and a 14.15\% average decrease. Removing video filter causes sharp degradation in multi-hop scenarios—up to 26.92\% in cross-spatio-temporal tasks—highlighting its importance for reducing irrelevant search space. Limiting tree depth results in a 6.39\% average drop, showing detailed lower-layer information significantly contributes to fine-grained reasoning.
Notably, in the \textit{w/o Video Filter} setting, downstream agents occasionally still select relevant clips through natural language understanding, implying fault tolerance and adaptivity in the multi-agent architecture.

\section{Conclusion}
This paper introduces VideoForest, a novel hierarchical framework for cross-video question answering that addresses the fundamental challenge of integrating and reasoning about information distributed across multiple video sources. By anchoring on person-level features as natural bridge points between videos, we enable sophisticated cross-video reasoning without requiring end-to-end training on multi-video datasets.
We developed CrossVideoQA, a comprehensive benchmark specifically designed for person-centric cross-video analysis, and demonstrated VideoForest's performance advantages over state-of-the-art models across single-video to cross-spatiotemporal configurations.
Our tree-based architecture with multi-agent reasoning establishes a foundation for cross-video understanding that overcomes isolated video processing limitations while maintaining computational tractability for real-world applications.

\begin{acks}
This work was supported by the National Natural Science Foundation of China (No. 62176223, 62302535, 62371305) and in part by
Guangdong Basic and Applied Basic Research Foundation (2023A1515011639, 2024A1515030025).
\end{acks}
\bibliographystyle{ACM-Reference-Format}
\balance
\bibliography{sample-base}


\begin{thebibliography}{50}


\ifx \showCODEN    \undefined \def \showCODEN     #1{\unskip}     \fi
\ifx \showISBNx    \undefined \def \showISBNx     #1{\unskip}     \fi
\ifx \showISBNxiii \undefined \def \showISBNxiii  #1{\unskip}     \fi
\ifx \showISSN     \undefined \def \showISSN      #1{\unskip}     \fi
\ifx \showLCCN     \undefined \def \showLCCN      #1{\unskip}     \fi
\ifx \shownote     \undefined \def \shownote      #1{#1}          \fi
\ifx \showarticletitle \undefined \def \showarticletitle #1{#1}   \fi
\ifx \showURL      \undefined \def \showURL       {\relax}        \fi
\providecommand\bibfield[2]{#2}
\providecommand\bibinfo[2]{#2}
\providecommand\natexlab[1]{#1}
\providecommand\showeprint[2][]{arXiv:#2}

\bibitem[Ben-Shabat et~al\mbox{.}(2023)]%
        {benshabat2023ikeaasmdatasetunderstanding}
\bibfield{author}{\bibinfo{person}{Yizhak Ben-Shabat}, \bibinfo{person}{Xin Yu}, \bibinfo{person}{Fatemeh~Sadat Saleh}, \bibinfo{person}{Dylan Campbell}, \bibinfo{person}{Cristian Rodriguez-Opazo}, \bibinfo{person}{Hongdong Li}, {and} \bibinfo{person}{Stephen Gould}.} \bibinfo{year}{2023}\natexlab{}.
\newblock \bibinfo{title}{The IKEA ASM Dataset: Understanding People Assembling Furniture through Actions, Objects and Pose}.
\newblock
\showeprint[arxiv]{2007.00394}~[cs.CV]
\urldef\tempurl%
\url{https://arxiv.org/abs/2007.00394}
\showURL{%
\tempurl}


\bibitem[Boqiang~Zhang(2025)]%
        {damonlpsg2025videollama3}
\bibfield{author}{\bibinfo{person}{Zesen Cheng-Zhiqiang Hu Yuqian Yuan Guanzheng Chen Sicong Leng Yuming Jiang Hang Zhang Xin Li Peng Jin Wenqi Zhang Fan Wang Lidong Bing Deli~Zhao Boqiang~Zhang, Kehan~Li}.} \bibinfo{year}{2025}\natexlab{}.
\newblock \showarticletitle{VideoLLaMA 3: Frontier Multimodal Foundation Models for Image and Video Understanding}.
\newblock \bibinfo{journal}{\emph{arXiv preprint arXiv:2501.13106}} (\bibinfo{year}{2025}).
\newblock
\urldef\tempurl%
\url{https://arxiv.org/abs/2501.13106}
\showURL{%
\tempurl}


\bibitem[Carreira et~al\mbox{.}(2022)]%
        {carreira2022shortnotekinetics700human}
\bibfield{author}{\bibinfo{person}{Joao Carreira}, \bibinfo{person}{Eric Noland}, \bibinfo{person}{Chloe Hillier}, {and} \bibinfo{person}{Andrew Zisserman}.} \bibinfo{year}{2022}\natexlab{}.
\newblock \bibinfo{title}{A Short Note on the Kinetics-700 Human Action Dataset}.
\newblock
\showeprint[arxiv]{1907.06987}~[cs.CV]
\urldef\tempurl%
\url{https://arxiv.org/abs/1907.06987}
\showURL{%
\tempurl}


\bibitem[Chen et~al\mbox{.}(2024)]%
        {chen2024sharegpt4video}
\bibfield{author}{\bibinfo{person}{Lin Chen}, \bibinfo{person}{Xilin Wei}, \bibinfo{person}{Jinsong Li}, \bibinfo{person}{Xiaoyi Dong}, \bibinfo{person}{Pan Zhang}, \bibinfo{person}{Yuhang Zang}, \bibinfo{person}{Zehui Chen}, \bibinfo{person}{Haodong Duan}, \bibinfo{person}{Bin Lin}, \bibinfo{person}{Zhenyu Tang}, \bibinfo{person}{Li Yuan}, \bibinfo{person}{Yu Qiao}, \bibinfo{person}{Dahua Lin}, \bibinfo{person}{Feng Zhao}, {and} \bibinfo{person}{Jiaqi Wang}.} \bibinfo{year}{2024}\natexlab{}.
\newblock \showarticletitle{ShareGPT4Video: Improving Video Understanding and Generation with Better Captions}.
\newblock \bibinfo{journal}{\emph{arXiv preprint arXiv:2406.04325}} (\bibinfo{year}{2024}).
\newblock


\bibitem[Cheng et~al\mbox{.}(2024b)]%
        {cheng2024enhancinglongvideounderstanding}
\bibfield{author}{\bibinfo{person}{Dingxin Cheng}, \bibinfo{person}{Mingda Li}, \bibinfo{person}{Jingyu Liu}, \bibinfo{person}{Yongxin Guo}, \bibinfo{person}{Bin Jiang}, \bibinfo{person}{Qingbin Liu}, \bibinfo{person}{Xi Chen}, {and} \bibinfo{person}{Bo Zhao}.} \bibinfo{year}{2024}\natexlab{b}.
\newblock \bibinfo{title}{Enhancing Long Video Understanding via Hierarchical Event-Based Memory}.
\newblock
\showeprint[arxiv]{2409.06299}~[cs.CV]
\urldef\tempurl%
\url{https://arxiv.org/abs/2409.06299}
\showURL{%
\tempurl}


\bibitem[Cheng et~al\mbox{.}(2024a)]%
        {damonlpsg2024videollama2}
\bibfield{author}{\bibinfo{person}{Zesen Cheng}, \bibinfo{person}{Sicong Leng}, \bibinfo{person}{Hang Zhang}, \bibinfo{person}{Yifei Xin}, \bibinfo{person}{Xin Li}, \bibinfo{person}{Guanzheng Chen}, \bibinfo{person}{Yongxin Zhu}, \bibinfo{person}{Wenqi Zhang}, \bibinfo{person}{Ziyang Luo}, \bibinfo{person}{Deli Zhao}, {and} \bibinfo{person}{Lidong Bing}.} \bibinfo{year}{2024}\natexlab{a}.
\newblock \showarticletitle{VideoLLaMA 2: Advancing Spatial-Temporal Modeling and Audio Understanding in Video-LLMs}.
\newblock \bibinfo{journal}{\emph{arXiv preprint arXiv:2406.07476}} (\bibinfo{year}{2024}).
\newblock
\urldef\tempurl%
\url{https://arxiv.org/abs/2406.07476}
\showURL{%
\tempurl}


\bibitem[Fan et~al\mbox{.}(2025)]%
        {fan2025videoagent}
\bibfield{author}{\bibinfo{person}{Yue Fan}, \bibinfo{person}{Xiaojian Ma}, \bibinfo{person}{Rujie Wu}, \bibinfo{person}{Yuntao Du}, \bibinfo{person}{Jiaqi Li}, \bibinfo{person}{Zhi Gao}, {and} \bibinfo{person}{Qing Li}.} \bibinfo{year}{2025}\natexlab{}.
\newblock \showarticletitle{Videoagent: A memory-augmented multimodal agent for video understanding}. In \bibinfo{booktitle}{\emph{European Conference on Computer Vision}}. Springer, \bibinfo{pages}{75--92}.
\newblock


\bibitem[Fei et~al\mbox{.}(2024b)]%
        {10508488}
\bibfield{author}{\bibinfo{person}{Hao Fei}, \bibinfo{person}{Shengqiong Wu}, \bibinfo{person}{Meishan Zhang}, \bibinfo{person}{Min Zhang}, \bibinfo{person}{Tat-Seng Chua}, {and} \bibinfo{person}{Shuicheng Yan}.} \bibinfo{year}{2024}\natexlab{b}.
\newblock \showarticletitle{Enhancing Video-Language Representations With Structural Spatio-Temporal Alignment}.
\newblock \bibinfo{journal}{\emph{IEEE Transactions on Pattern Analysis and Machine Intelligence}} \bibinfo{volume}{46}, \bibinfo{number}{12} (\bibinfo{year}{2024}), \bibinfo{pages}{7701--7719}.
\newblock
\href{https://doi.org/10.1109/TPAMI.2024.3393452}{doi:\nolinkurl{10.1109/TPAMI.2024.3393452}}


\bibitem[Fei et~al\mbox{.}(2024a)]%
        {fei2024videoccamenhancingvideolanguageunderstanding}
\bibfield{author}{\bibinfo{person}{Jiajun Fei}, \bibinfo{person}{Dian Li}, \bibinfo{person}{Zhidong Deng}, \bibinfo{person}{Zekun Wang}, \bibinfo{person}{Gang Liu}, {and} \bibinfo{person}{Hui Wang}.} \bibinfo{year}{2024}\natexlab{a}.
\newblock \bibinfo{title}{Video-CCAM: Enhancing Video-Language Understanding with Causal Cross-Attention Masks for Short and Long Videos}.
\newblock
\showeprint[arxiv]{2408.14023}~[cs.CV]
\urldef\tempurl%
\url{https://arxiv.org/abs/2408.14023}
\showURL{%
\tempurl}


\bibitem[Fu et~al\mbox{.}(2024)]%
        {fu2024videoMMEBenchmark}
\bibfield{author}{\bibinfo{person}{Chaoyou Fu}, \bibinfo{person}{Yuhan Dai}, \bibinfo{person}{Yondong Luo}, \bibinfo{person}{Lei Li}, \bibinfo{person}{Shuhuai Ren}, \bibinfo{person}{Renrui Zhang}, \bibinfo{person}{Zihan Wang}, \bibinfo{person}{Chenyu Zhou}, \bibinfo{person}{Yunhang Shen}, \bibinfo{person}{Mengdan Zhang}, {et~al\mbox{.}}} \bibinfo{year}{2024}\natexlab{}.
\newblock \showarticletitle{Video-MME: The First-Ever Comprehensive Evaluation Benchmark of Multi-modal LLMs in Video Analysis}.
\newblock \bibinfo{journal}{\emph{arXiv preprint arXiv:2405.21075}} (\bibinfo{year}{2024}).
\newblock


\bibitem[Heilbron et~al\mbox{.}(2015)]%
        {Heilbron2015ActivityNetAL}
\bibfield{author}{\bibinfo{person}{Fabian~Caba Heilbron}, \bibinfo{person}{Victor Escorcia}, \bibinfo{person}{Bernard Ghanem}, {and} \bibinfo{person}{Juan~Carlos Niebles}.} \bibinfo{year}{2015}\natexlab{}.
\newblock \showarticletitle{ActivityNet: A large-scale video benchmark for human activity understanding}.
\newblock \bibinfo{journal}{\emph{2015 IEEE Conference on Computer Vision and Pattern Recognition (CVPR)}} (\bibinfo{year}{2015}), \bibinfo{pages}{961--970}.
\newblock
\urldef\tempurl%
\url{https://api.semanticscholar.org/CorpusID:1710722}
\showURL{%
\tempurl}


\bibitem[Islam et~al\mbox{.}(2024)]%
        {islam2024video}
\bibfield{author}{\bibinfo{person}{Md~Mohaiminul Islam}, \bibinfo{person}{Ngan Ho}, \bibinfo{person}{Xitong Yang}, \bibinfo{person}{Tushar Nagarajan}, \bibinfo{person}{Lorenzo Torresani}, {and} \bibinfo{person}{Gedas Bertasius}.} \bibinfo{year}{2024}\natexlab{}.
\newblock \showarticletitle{Video ReCap: Recursive Captioning of Hour-Long Videos}.
\newblock \bibinfo{journal}{\emph{arXiv preprint arXiv:2402.13250}} (\bibinfo{year}{2024}).
\newblock


\bibitem[Islam et~al\mbox{.}(2025)]%
        {islam2025bimba}
\bibfield{author}{\bibinfo{person}{Md~Mohaiminul Islam}, \bibinfo{person}{Tushar Nagarajan}, \bibinfo{person}{Huiyu Wang}, \bibinfo{person}{Gedas Bertasius}, {and} \bibinfo{person}{Lorenzo Torresani}.} \bibinfo{year}{2025}\natexlab{}.
\newblock \showarticletitle{BIMBA: Selective-Scan Compression for Long-Range Video Question Answering}.
\newblock \bibinfo{journal}{\emph{arXiv preprint arXiv:2503.09590}} (\bibinfo{year}{2025}).
\newblock


\bibitem[Labs(2024)]%
        {crewai2024}
\bibfield{author}{\bibinfo{person}{Cognition Labs}.} \bibinfo{year}{2024}\natexlab{}.
\newblock \bibinfo{title}{CrewAI: Framework for Building Multi-Agent Systems}.
\newblock \bibinfo{howpublished}{\url{https://github.com/joaomdmoura/crewai}}.
\newblock
\newblock
\shownote{Accessed: 2025-04}.


\bibitem[Li et~al\mbox{.}(2024d)]%
        {li2024llavaonevision}
\bibfield{author}{\bibinfo{person}{Bo Li}, \bibinfo{person}{Yuanhan Zhang}, \bibinfo{person}{Dong Guo}, \bibinfo{person}{Renrui Zhang}, \bibinfo{person}{Feng Li}, \bibinfo{person}{Hao Zhang}, \bibinfo{person}{Kaichen Zhang}, \bibinfo{person}{Yanwei Li}, \bibinfo{person}{Ziwei Liu}, {and} \bibinfo{person}{Chunyuan Li}.} \bibinfo{year}{2024}\natexlab{d}.
\newblock \showarticletitle{LLaVA-OneVision: Easy Visual Task Transfer}.
\newblock \bibinfo{journal}{\emph{arXiv preprint arXiv:2408.03326}} (\bibinfo{year}{2024}).
\newblock


\bibitem[Li et~al\mbox{.}(2024b)]%
        {li2024mvbench}
\bibfield{author}{\bibinfo{person}{Kunchang Li}, \bibinfo{person}{Yali Wang}, \bibinfo{person}{Yinan He}, \bibinfo{person}{Yizhuo Li}, \bibinfo{person}{Yi Wang}, \bibinfo{person}{Yi Liu}, \bibinfo{person}{Zun Wang}, \bibinfo{person}{Jilan Xu}, \bibinfo{person}{Guo Chen}, \bibinfo{person}{Ping Luo}, {et~al\mbox{.}}} \bibinfo{year}{2024}\natexlab{b}.
\newblock \showarticletitle{Mvbench: A comprehensive multi-modal video understanding benchmark}. In \bibinfo{booktitle}{\emph{Proceedings of the IEEE/CVF Conference on Computer Vision and Pattern Recognition}}. \bibinfo{pages}{22195--22206}.
\newblock


\bibitem[Li et~al\mbox{.}(2023)]%
        {10.1145/3581783.3612838}
\bibfield{author}{\bibinfo{person}{Ruizhe Li}, \bibinfo{person}{Jiahao Guo}, \bibinfo{person}{Mingxi Li}, \bibinfo{person}{Zhengqian Wu}, {and} \bibinfo{person}{Chao Liang}.} \bibinfo{year}{2023}\natexlab{}.
\newblock \showarticletitle{A Hierarchical Deep Video Understanding Method with Shot-Based Instance Search and Large Language Model}. In \bibinfo{booktitle}{\emph{Proceedings of the 31st ACM International Conference on Multimedia}} (Ottawa ON, Canada) \emph{(\bibinfo{series}{MM '23})}. \bibinfo{publisher}{Association for Computing Machinery}, \bibinfo{address}{New York, NY, USA}, \bibinfo{pages}{9425–9429}.
\newblock
\showISBNx{9798400701085}
\href{https://doi.org/10.1145/3581783.3612838}{doi:\nolinkurl{10.1145/3581783.3612838}}


\bibitem[Li et~al\mbox{.}(2024c)]%
        {li2024videochatflash}
\bibfield{author}{\bibinfo{person}{Xinhao Li}, \bibinfo{person}{Yi Wang}, \bibinfo{person}{Jiashuo Yu}, \bibinfo{person}{Xiangyu Zeng}, \bibinfo{person}{Yuhan Zhu}, \bibinfo{person}{Haian Huang}, \bibinfo{person}{Jianfei Gao}, \bibinfo{person}{Kunchang Li}, \bibinfo{person}{Yinan He}, \bibinfo{person}{Chenting Wang}, {et~al\mbox{.}}} \bibinfo{year}{2024}\natexlab{c}.
\newblock \showarticletitle{VideoChat-Flash: Hierarchical Compression for Long-Context Video Modeling}.
\newblock \bibinfo{journal}{\emph{arXiv preprint arXiv:2501.00574}} (\bibinfo{year}{2024}).
\newblock


\bibitem[Li et~al\mbox{.}(2024a)]%
        {Li2024VideoVistaAV}
\bibfield{author}{\bibinfo{person}{Yunxin Li}, \bibinfo{person}{Xinyu Chen}, \bibinfo{person}{Baotian Hu}, \bibinfo{person}{Longyue Wang}, \bibinfo{person}{Haoyuan Shi}, {and} \bibinfo{person}{Min Zhang}.} \bibinfo{year}{2024}\natexlab{a}.
\newblock \showarticletitle{VideoVista: A Versatile Benchmark for Video Understanding and Reasoning}.
\newblock \bibinfo{journal}{\emph{ArXiv}}  \bibinfo{volume}{abs/2406.11303} (\bibinfo{year}{2024}).
\newblock
\urldef\tempurl%
\url{https://api.semanticscholar.org/CorpusID:270559556}
\showURL{%
\tempurl}


\bibitem[Liu et~al\mbox{.}(2025a)]%
        {liu2025commonsensevideoquestionanswering}
\bibfield{author}{\bibinfo{person}{Huabin Liu}, \bibinfo{person}{Filip Ilievski}, {and} \bibinfo{person}{Cees G.~M. Snoek}.} \bibinfo{year}{2025}\natexlab{a}.
\newblock \bibinfo{title}{Commonsense Video Question Answering through Video-Grounded Entailment Tree Reasoning}.
\newblock
\showeprint[arxiv]{2501.05069}~[cs.CV]
\urldef\tempurl%
\url{https://arxiv.org/abs/2501.05069}
\showURL{%
\tempurl}


\bibitem[Liu et~al\mbox{.}(2024a)]%
        {liu2024llavanext}
\bibfield{author}{\bibinfo{person}{Haotian Liu}, \bibinfo{person}{Chunyuan Li}, \bibinfo{person}{Yuheng Li}, \bibinfo{person}{Bo Li}, \bibinfo{person}{Yuanhan Zhang}, \bibinfo{person}{Sheng Shen}, {and} \bibinfo{person}{Yong~Jae Lee}.} \bibinfo{year}{2024}\natexlab{a}.
\newblock \bibinfo{title}{LLaVA-NeXT: Improved reasoning, OCR, and world knowledge}.
\newblock
\urldef\tempurl%
\url{https://llava-vl.github.io/blog/2024-01-30-llava-next/}
\showURL{%
\tempurl}


\bibitem[Liu et~al\mbox{.}(2024b)]%
        {10.1007/978-3-031-72652-1_6}
\bibfield{author}{\bibinfo{person}{Huabin Liu}, \bibinfo{person}{Xiao Ma}, \bibinfo{person}{Cheng Zhong}, \bibinfo{person}{Yang Zhang}, {and} \bibinfo{person}{Weiyao Lin}.} \bibinfo{year}{2024}\natexlab{b}.
\newblock \showarticletitle{TimeCraft: Navigate Weakly-Supervised Temporal Grounded Video Question Answering via Bi-directional Reasoning}. In \bibinfo{booktitle}{\emph{Computer Vision – ECCV 2024: 18th European Conference, Milan, Italy, September 29–October 4, 2024, Proceedings, Part V}} (Milan, Italy). \bibinfo{publisher}{Springer-Verlag}, \bibinfo{address}{Berlin, Heidelberg}, \bibinfo{pages}{92–107}.
\newblock
\showISBNx{978-3-031-72651-4}
\href{https://doi.org/10.1007/978-3-031-72652-1_6}{doi:\nolinkurl{10.1007/978-3-031-72652-1_6}}


\bibitem[Liu et~al\mbox{.}(2025b)]%
        {liu2025semantic}
\bibfield{author}{\bibinfo{person}{Mingyang Liu}, \bibinfo{person}{Fan Zhou}, \bibinfo{person}{Ruomei Wang}, \bibinfo{person}{Baoquan Zhao}, {and} \bibinfo{person}{Fuwei Zhang}.} \bibinfo{year}{2025}\natexlab{b}.
\newblock \showarticletitle{Semantic Distance-Aware Cross-Modal Attention Mechanism for Video Question Answering}.
\newblock \bibinfo{journal}{\emph{IEEE Transactions on Instrumentation and Measurement}} (\bibinfo{year}{2025}).
\newblock


\bibitem[Liu et~al\mbox{.}(2022)]%
        {liu2022fineactionfinegrainedvideodataset}
\bibfield{author}{\bibinfo{person}{Yi Liu}, \bibinfo{person}{Limin Wang}, \bibinfo{person}{Yali Wang}, \bibinfo{person}{Xiao Ma}, {and} \bibinfo{person}{Yu Qiao}.} \bibinfo{year}{2022}\natexlab{}.
\newblock \bibinfo{title}{FineAction: A Fine-Grained Video Dataset for Temporal Action Localization}.
\newblock
\showeprint[arxiv]{2105.11107}~[cs.CV]
\urldef\tempurl%
\url{https://arxiv.org/abs/2105.11107}
\showURL{%
\tempurl}


\bibitem[Ma et~al\mbox{.}(2024)]%
        {ma2024drvideodocumentretrievalbased}
\bibfield{author}{\bibinfo{person}{Ziyu Ma}, \bibinfo{person}{Chenhui Gou}, \bibinfo{person}{Hengcan Shi}, \bibinfo{person}{Bin Sun}, \bibinfo{person}{Shutao Li}, \bibinfo{person}{Hamid Rezatofighi}, {and} \bibinfo{person}{Jianfei Cai}.} \bibinfo{year}{2024}\natexlab{}.
\newblock \bibinfo{title}{DrVideo: Document Retrieval Based Long Video Understanding}.
\newblock
\showeprint[arxiv]{2406.12846}~[cs.CV]
\urldef\tempurl%
\url{https://arxiv.org/abs/2406.12846}
\showURL{%
\tempurl}


\bibitem[Maaz et~al\mbox{.}(2024)]%
        {Maaz2023VideoChatGPT}
\bibfield{author}{\bibinfo{person}{Muhammad Maaz}, \bibinfo{person}{Hanoona Rasheed}, \bibinfo{person}{Salman Khan}, {and} \bibinfo{person}{Fahad~Shahbaz Khan}.} \bibinfo{year}{2024}\natexlab{}.
\newblock \showarticletitle{Video-ChatGPT: Towards Detailed Video Understanding via Large Vision and Language Models}. In \bibinfo{booktitle}{\emph{Proceedings of the 62nd Annual Meeting of the Association for Computational Linguistics (ACL 2024)}}.
\newblock


\bibitem[Madan et~al\mbox{.}(2024)]%
        {madan2024foundationmodelsvideounderstanding}
\bibfield{author}{\bibinfo{person}{Neelu Madan}, \bibinfo{person}{Andreas Moegelmose}, \bibinfo{person}{Rajat Modi}, \bibinfo{person}{Yogesh~S. Rawat}, {and} \bibinfo{person}{Thomas~B. Moeslund}.} \bibinfo{year}{2024}\natexlab{}.
\newblock \bibinfo{title}{Foundation Models for Video Understanding: A Survey}.
\newblock
\showeprint[arxiv]{2405.03770}~[cs.CV]
\urldef\tempurl%
\url{https://arxiv.org/abs/2405.03770}
\showURL{%
\tempurl}


\bibitem[Mangalam et~al\mbox{.}(2023)]%
        {mangalam2023egoschema}
\bibfield{author}{\bibinfo{person}{Karttikeya Mangalam}, \bibinfo{person}{Raiymbek Akshulakov}, {and} \bibinfo{person}{Jitendra Malik}.} \bibinfo{year}{2023}\natexlab{}.
\newblock \showarticletitle{Egoschema: A diagnostic benchmark for very long-form video language understanding}.
\newblock \bibinfo{journal}{\emph{Advances in Neural Information Processing Systems}}  \bibinfo{volume}{36} (\bibinfo{year}{2023}), \bibinfo{pages}{46212--46244}.
\newblock


\bibitem[Ning et~al\mbox{.}(2023)]%
        {ning2023video}
\bibfield{author}{\bibinfo{person}{Munan Ning}, \bibinfo{person}{Bin Zhu}, \bibinfo{person}{Yujia Xie}, \bibinfo{person}{Bin Lin}, \bibinfo{person}{Jiaxi Cui}, \bibinfo{person}{Lu Yuan}, \bibinfo{person}{Dongdong Chen}, {and} \bibinfo{person}{Li Yuan}.} \bibinfo{year}{2023}\natexlab{}.
\newblock \showarticletitle{Video-bench: A comprehensive benchmark and toolkit for evaluating video-based large language models}.
\newblock \bibinfo{journal}{\emph{arXiv preprint arXiv:2311.16103}} (\bibinfo{year}{2023}).
\newblock


\bibitem[Park et~al\mbox{.}(2024)]%
        {Park2024TooMF}
\bibfield{author}{\bibinfo{person}{Jong~Sung Park}, \bibinfo{person}{Kanchana Ranasinghe}, \bibinfo{person}{Kumara Kahatapitiya}, \bibinfo{person}{Wonjeong Ryoo}, \bibinfo{person}{Donghyun Kim}, {and} \bibinfo{person}{Michael~S. Ryoo}.} \bibinfo{year}{2024}\natexlab{}.
\newblock \showarticletitle{Too Many Frames, not all Useful: Efficient Strategies for Long-Form Video QA}.
\newblock \bibinfo{journal}{\emph{ArXiv}}  \bibinfo{volume}{abs/2406.09396} (\bibinfo{year}{2024}).
\newblock
\urldef\tempurl%
\url{https://api.semanticscholar.org/CorpusID:270440923}
\showURL{%
\tempurl}


\bibitem[Qasim et~al\mbox{.}(2021)]%
        {qasim2021groundtruthing}
\bibfield{author}{\bibinfo{person}{T. Qasim}, \bibinfo{person}{R.~B. Fisher}, {and} \bibinfo{person}{N. Bhatti}.} \bibinfo{year}{2021}\natexlab{}.
\newblock \showarticletitle{Ground-truthing Large Human Behavior Monitoring Datasets}. In \bibinfo{booktitle}{\emph{Proceedings of the 2020 International Conference on Pattern Recognition}}. \bibinfo{address}{Online}.
\newblock


\bibitem[Ren et~al\mbox{.}(2024)]%
        {10656135}
\bibfield{author}{\bibinfo{person}{Shuhuai Ren}, \bibinfo{person}{Linli Yao}, \bibinfo{person}{Shicheng Li}, \bibinfo{person}{Xu Sun}, {and} \bibinfo{person}{Lu Hou}.} \bibinfo{year}{2024}\natexlab{}.
\newblock \showarticletitle{TimeChat: A Time-sensitive Multimodal Large Language Model for Long Video Understanding}. In \bibinfo{booktitle}{\emph{2024 IEEE/CVF Conference on Computer Vision and Pattern Recognition (CVPR)}}. \bibinfo{pages}{14313--14323}.
\newblock
\href{https://doi.org/10.1109/CVPR52733.2024.01357}{doi:\nolinkurl{10.1109/CVPR52733.2024.01357}}


\bibitem[Sanders et~al\mbox{.}(2024)]%
        {sanders2024tvtreesmultimodalentailmenttrees}
\bibfield{author}{\bibinfo{person}{Kate Sanders}, \bibinfo{person}{Nathaniel Weir}, {and} \bibinfo{person}{Benjamin~Van Durme}.} \bibinfo{year}{2024}\natexlab{}.
\newblock \bibinfo{title}{TV-TREES: Multimodal Entailment Trees for Neuro-Symbolic Video Reasoning}.
\newblock
\showeprint[arxiv]{2402.19467}~[cs.CL]
\urldef\tempurl%
\url{https://arxiv.org/abs/2402.19467}
\showURL{%
\tempurl}


\bibitem[Sigurdsson et~al\mbox{.}(2016)]%
        {sigurdsson2016hollywoodhomescrowdsourcingdata}
\bibfield{author}{\bibinfo{person}{Gunnar~A. Sigurdsson}, \bibinfo{person}{Gül Varol}, \bibinfo{person}{Xiaolong Wang}, \bibinfo{person}{Ali Farhadi}, \bibinfo{person}{Ivan Laptev}, {and} \bibinfo{person}{Abhinav Gupta}.} \bibinfo{year}{2016}\natexlab{}.
\newblock \bibinfo{title}{Hollywood in Homes: Crowdsourcing Data Collection for Activity Understanding}.
\newblock
\showeprint[arxiv]{1604.01753}~[cs.CV]
\urldef\tempurl%
\url{https://arxiv.org/abs/1604.01753}
\showURL{%
\tempurl}


\bibitem[Wang et~al\mbox{.}(2023a)]%
        {wang2023chatvideo}
\bibfield{author}{\bibinfo{person}{Junke Wang}, \bibinfo{person}{Dongdong Chen}, \bibinfo{person}{Chong Luo}, \bibinfo{person}{Xiyang Dai}, \bibinfo{person}{Lu Yuan}, \bibinfo{person}{Zuxuan Wu}, {and} \bibinfo{person}{Yu-Gang Jiang}.} \bibinfo{year}{2023}\natexlab{a}.
\newblock \showarticletitle{Chatvideo: A tracklet-centric multimodal and versatile video understanding system}.
\newblock \bibinfo{journal}{\emph{arXiv preprint arXiv:2304.14407}} (\bibinfo{year}{2023}).
\newblock


\bibitem[Wang et~al\mbox{.}(2023b)]%
        {wang2023internvid}
\bibfield{author}{\bibinfo{person}{Yi Wang}, \bibinfo{person}{Yinan He}, \bibinfo{person}{Yizhuo Li}, \bibinfo{person}{Kunchang Li}, \bibinfo{person}{Jiashuo Yu}, \bibinfo{person}{Xin Ma}, \bibinfo{person}{Xinhao Li}, \bibinfo{person}{Guo Chen}, \bibinfo{person}{Xinyuan Chen}, \bibinfo{person}{Yaohui Wang}, {et~al\mbox{.}}} \bibinfo{year}{2023}\natexlab{b}.
\newblock \showarticletitle{InternVid: A Large-scale Video-Text Dataset for Multimodal Understanding and Generation}. In \bibinfo{booktitle}{\emph{The Twelfth International Conference on Learning Representations}}.
\newblock


\bibitem[Wang et~al\mbox{.}(2022)]%
        {wang2022internvideo}
\bibfield{author}{\bibinfo{person}{Yi Wang}, \bibinfo{person}{Kunchang Li}, \bibinfo{person}{Yizhuo Li}, \bibinfo{person}{Yinan He}, \bibinfo{person}{Bingkun Huang}, \bibinfo{person}{Zhiyu Zhao}, \bibinfo{person}{Hongjie Zhang}, \bibinfo{person}{Jilan Xu}, \bibinfo{person}{Yi Liu}, \bibinfo{person}{Zun Wang}, \bibinfo{person}{Sen Xing}, \bibinfo{person}{Guo Chen}, \bibinfo{person}{Junting Pan}, \bibinfo{person}{Jiashuo Yu}, \bibinfo{person}{Yali Wang}, \bibinfo{person}{Limin Wang}, {and} \bibinfo{person}{Yu Qiao}.} \bibinfo{year}{2022}\natexlab{}.
\newblock \showarticletitle{InternVideo: General Video Foundation Models via Generative and Discriminative Learning}.
\newblock \bibinfo{journal}{\emph{arXiv preprint arXiv:2212.03191}} (\bibinfo{year}{2022}).
\newblock


\bibitem[Wang et~al\mbox{.}(2025)]%
        {wang2025internvideo}
\bibfield{author}{\bibinfo{person}{Yi Wang}, \bibinfo{person}{Xinhao Li}, \bibinfo{person}{Ziang Yan}, \bibinfo{person}{Yinan He}, \bibinfo{person}{Jiashuo Yu}, \bibinfo{person}{Xiangyu Zeng}, \bibinfo{person}{Chenting Wang}, \bibinfo{person}{Changlian Ma}, \bibinfo{person}{Haian Huang}, \bibinfo{person}{Jianfei Gao}, \bibinfo{person}{Min Dou}, \bibinfo{person}{Kai Chen}, \bibinfo{person}{Wenhai Wang}, \bibinfo{person}{Yu Qiao}, \bibinfo{person}{Yali Wang}, {and} \bibinfo{person}{Limin Wang}.} \bibinfo{year}{2025}\natexlab{}.
\newblock \showarticletitle{InternVideo2.5: Empowering Video MLLMs with Long and Rich Context Modeling}.
\newblock \bibinfo{journal}{\emph{arXiv preprint arXiv:2501.12386}} (\bibinfo{year}{2025}).
\newblock


\bibitem[Wang et~al\mbox{.}(2021)]%
        {Wang_2021_ICCV}
\bibfield{author}{\bibinfo{person}{Yingquan Wang}, \bibinfo{person}{Pingping Zhang}, \bibinfo{person}{Shang Gao}, \bibinfo{person}{Xia Geng}, \bibinfo{person}{Hu Lu}, {and} \bibinfo{person}{Dong Wang}.} \bibinfo{year}{2021}\natexlab{}.
\newblock \showarticletitle{Pyramid Spatial-Temporal Aggregation for Video-Based Person Re-Identification}. In \bibinfo{booktitle}{\emph{Proceedings of the IEEE/CVF International Conference on Computer Vision (ICCV)}}. \bibinfo{pages}{12026--12035}.
\newblock


\bibitem[Wu et~al\mbox{.}(2024)]%
        {wu2024longvideobenchbenchmarklongcontextinterleaved}
\bibfield{author}{\bibinfo{person}{Haoning Wu}, \bibinfo{person}{Dongxu Li}, \bibinfo{person}{Bei Chen}, {and} \bibinfo{person}{Junnan Li}.} \bibinfo{year}{2024}\natexlab{}.
\newblock \bibinfo{title}{LongVideoBench: A Benchmark for Long-context Interleaved Video-Language Understanding}.
\newblock
\showeprint[arxiv]{2407.15754}~[cs.CV]
\urldef\tempurl%
\url{https://arxiv.org/abs/2407.15754}
\showURL{%
\tempurl}


\bibitem[Xiao et~al\mbox{.}(2022)]%
        {xiao2022video}
\bibfield{author}{\bibinfo{person}{Junbin Xiao}, \bibinfo{person}{Angela Yao}, \bibinfo{person}{Zhiyuan Liu}, \bibinfo{person}{Yicong Li}, \bibinfo{person}{Wei Ji}, {and} \bibinfo{person}{Tat-Seng Chua}.} \bibinfo{year}{2022}\natexlab{}.
\newblock \showarticletitle{Video as conditional graph hierarchy for multi-granular question answering}. In \bibinfo{booktitle}{\emph{Proceedings of the AAAI Conference on Artificial Intelligence}}, Vol.~\bibinfo{volume}{36}. \bibinfo{pages}{2804--2812}.
\newblock


\bibitem[Xiong et~al\mbox{.}(2024)]%
        {xiong2024llavacritic}
\bibfield{author}{\bibinfo{person}{Tianyi Xiong}, \bibinfo{person}{Xiyao Wang}, \bibinfo{person}{Dong Guo}, \bibinfo{person}{Qinghao Ye}, \bibinfo{person}{Haoqi Fan}, \bibinfo{person}{Quanquan Gu}, \bibinfo{person}{Heng Huang}, {and} \bibinfo{person}{Chunyuan Li}.} \bibinfo{year}{2024}\natexlab{}.
\newblock \showarticletitle{LLaVA-Critic: Learning to Evaluate Multimodal Models}.
\newblock  (\bibinfo{year}{2024}).
\newblock
\showeprint[arxiv]{2410.02712}~[cs.CV]
\urldef\tempurl%
\url{https://arxiv.org/abs/2410.02712}
\showURL{%
\tempurl}


\bibitem[Ye et~al\mbox{.}(2024)]%
        {ye2024mplugowl3longimagesequenceunderstanding}
\bibfield{author}{\bibinfo{person}{Jiabo Ye}, \bibinfo{person}{Haiyang Xu}, \bibinfo{person}{Haowei Liu}, \bibinfo{person}{Anwen Hu}, \bibinfo{person}{Ming Yan}, \bibinfo{person}{Qi Qian}, \bibinfo{person}{Ji Zhang}, \bibinfo{person}{Fei Huang}, {and} \bibinfo{person}{Jingren Zhou}.} \bibinfo{year}{2024}\natexlab{}.
\newblock \bibinfo{title}{mPLUG-Owl3: Towards Long Image-Sequence Understanding in Multi-Modal Large Language Models}.
\newblock
\showeprint[arxiv]{2408.04840}~[cs.CV]
\urldef\tempurl%
\url{https://arxiv.org/abs/2408.04840}
\showURL{%
\tempurl}


\bibitem[Yu et~al\mbox{.}(2023)]%
        {yu2023selfchainedimagelanguagemodelvideo}
\bibfield{author}{\bibinfo{person}{Shoubin Yu}, \bibinfo{person}{Jaemin Cho}, \bibinfo{person}{Prateek Yadav}, {and} \bibinfo{person}{Mohit Bansal}.} \bibinfo{year}{2023}\natexlab{}.
\newblock \bibinfo{title}{Self-Chained Image-Language Model for Video Localization and Question Answering}.
\newblock
\showeprint[arxiv]{2305.06988}~[cs.CV]
\urldef\tempurl%
\url{https://arxiv.org/abs/2305.06988}
\showURL{%
\tempurl}


\bibitem[Zhang et~al\mbox{.}(2024b)]%
        {zhang2024simplellmframeworklongrange}
\bibfield{author}{\bibinfo{person}{Ce Zhang}, \bibinfo{person}{Taixi Lu}, \bibinfo{person}{Md~Mohaiminul Islam}, \bibinfo{person}{Ziyang Wang}, \bibinfo{person}{Shoubin Yu}, \bibinfo{person}{Mohit Bansal}, {and} \bibinfo{person}{Gedas Bertasius}.} \bibinfo{year}{2024}\natexlab{b}.
\newblock \bibinfo{title}{A Simple LLM Framework for Long-Range Video Question-Answering}.
\newblock
\showeprint[arxiv]{2312.17235}~[cs.CV]
\urldef\tempurl%
\url{https://arxiv.org/abs/2312.17235}
\showURL{%
\tempurl}


\bibitem[Zhang et~al\mbox{.}(2023)]%
        {damonlpsg2023videollama}
\bibfield{author}{\bibinfo{person}{Hang Zhang}, \bibinfo{person}{Xin Li}, {and} \bibinfo{person}{Lidong Bing}.} \bibinfo{year}{2023}\natexlab{}.
\newblock \showarticletitle{Video-LLaMA: An Instruction-tuned Audio-Visual Language Model for Video Understanding}.
\newblock \bibinfo{journal}{\emph{arXiv preprint arXiv:2306.02858}} (\bibinfo{year}{2023}).
\newblock
\urldef\tempurl%
\url{https://arxiv.org/abs/2306.02858}
\showURL{%
\tempurl}


\bibitem[Zhang et~al\mbox{.}(2024a)]%
        {zhang2024llavanextvideo}
\bibfield{author}{\bibinfo{person}{Yuanhan Zhang}, \bibinfo{person}{Bo Li}, \bibinfo{person}{haotian Liu}, \bibinfo{person}{Yong~jae Lee}, \bibinfo{person}{Liangke Gui}, \bibinfo{person}{Di Fu}, \bibinfo{person}{Jiashi Feng}, \bibinfo{person}{Ziwei Liu}, {and} \bibinfo{person}{Chunyuan Li}.} \bibinfo{year}{2024}\natexlab{a}.
\newblock \bibinfo{title}{LLaVA-NeXT: A Strong Zero-shot Video Understanding Model}.
\newblock
\urldef\tempurl%
\url{https://llava-vl.github.io/blog/2024-04-30-llava-next-video/}
\showURL{%
\tempurl}


\bibitem[Zhao et~al\mbox{.}(2019)]%
        {zhao2019hacs}
\bibfield{author}{\bibinfo{person}{Hang Zhao}, \bibinfo{person}{Zhicheng Yan}, \bibinfo{person}{Lorenzo Torresani}, {and} \bibinfo{person}{Antonio Torralba}.} \bibinfo{year}{2019}\natexlab{}.
\newblock \showarticletitle{HACS: Human Action Clips and Segments Dataset for Recognition and Temporal Localization}.
\newblock \bibinfo{journal}{\emph{arXiv preprint arXiv:1712.09374}} (\bibinfo{year}{2019}).
\newblock


\bibitem[Zhou et~al\mbox{.}(2023)]%
        {Zhou_2023}
\bibfield{author}{\bibinfo{person}{Qianyu Zhou}, \bibinfo{person}{Xiangtai Li}, \bibinfo{person}{Lu He}, \bibinfo{person}{Yibo Yang}, \bibinfo{person}{Guangliang Cheng}, \bibinfo{person}{Yunhai Tong}, \bibinfo{person}{Lizhuang Ma}, {and} \bibinfo{person}{Dacheng Tao}.} \bibinfo{year}{2023}\natexlab{}.
\newblock \showarticletitle{TransVOD: End-to-End Video Object Detection With Spatial-Temporal Transformers}.
\newblock \bibinfo{journal}{\emph{IEEE Transactions on Pattern Analysis and Machine Intelligence}} \bibinfo{volume}{45}, \bibinfo{number}{6} (\bibinfo{date}{June} \bibinfo{year}{2023}), \bibinfo{pages}{7853–7869}.
\newblock
\showISSN{1939-3539}
\href{https://doi.org/10.1109/tpami.2022.3223955}{doi:\nolinkurl{10.1109/tpami.2022.3223955}}


\bibitem[Ziyang~Wang(2024)]%
        {wang2024videotree}
\bibfield{author}{\bibinfo{person}{Elias Stengel-Eskin Jaehong Yoon Feng Cheng Gedas Bertasius Mohit~Bansal Ziyang~Wang, Shoubin~Yu}.} \bibinfo{year}{2024}\natexlab{}.
\newblock \showarticletitle{VideoTree: Adaptive Tree-based Video Representation for LLM Reasoning on Long Videos}.
\newblock \bibinfo{journal}{\emph{arxiv}} (\bibinfo{year}{2024}).
\newblock


\end{thebibliography}

\end{sloppypar}
\end{document}